%% file: sn-article.tex
\documentclass[sn-nature]{sn-jnl}% Style for submissions to Nature Portfolio journals
\usepackage{graphicx}%
\usepackage{multirow}%
\usepackage{amsmath,amssymb,amsfonts}%
\usepackage{amsthm}%
\usepackage{mathrsfs}%
\usepackage[title]{appendix}%
\usepackage{xcolor}%
\usepackage{textcomp}%
\usepackage{manyfoot}%
\usepackage{booktabs}%
\usepackage{algorithm}%
\usepackage{algorithmicx}%
\usepackage{algpseudocode}%
\usepackage{listings}%
\usepackage{cleveref}
\usepackage{tikz}
\newcommand\figref{Figure~\ref}

\newcommand\appref{Appendix~\ref}

\usepackage{caption}
\usepackage{subcaption}
\usepackage{xcolor}
\usepackage{wrapfig, blindtext}

\usepackage{algorithm}
\usepackage{algcompatible}

\begin{document}
% \backgroundsetup{contents={Please do not share}}

\title[Leveraging Cardiovascular Simulations for In-Vivo Prediction of Cardiac Biomarkers]{Leveraging Cardiovascular Simulations for In-Vivo Prediction of Cardiac Biomarkers}
%\title[Leveraging Cardiovascular Simulations for In-Vivo Prediction of Cardiac Biomarkers Using Simulation-Based Inference]{Leveraging Cardiovascular Simulations for In-Vivo Prediction of Cardiac Biomarkers Using Simulation-Based Inference}

% Alternative titles:
% Leveraging Cardiovascular Simulations for In-vivo Prediction of Cardiac Biomarkers
% Simulation-based Inference for In-vivo Prediction of Cardiac Biomarkers


\author[1]{\fnm{Laura} \sur{Manduchi}}
\equalcont{These authors contributed equally to this work.}
\author*[2]{\fnm{Antoine} \sur{Wehenkel}}\email{awehenkel@apple.com}

\equalcont{These authors contributed equally to this work.}

\author[2]{\fnm{Jens} \sur{Behrmann}}
\author[2]{\fnm{Luca} \sur{Pegolotti}}
\author[2]{\fnm{Andy} \sur{C.~Miller}}
\author[2]{\fnm{Marco} \sur{Cuturi}}
\author[2]{\fnm{Ozan} \sur{Sener}}
\author[2]{\fnm{Guillermo} \sur{Sapiro}}
\author[2]{\fnm{Jörn-Henrik} \sur{Jacobsen}}

\affil[1]{\orgdiv{ETH Zürich}}

\affil[2]{\orgname{Apple}}

% \affil[3]{\orgdiv{Department}, \orgname{Organization}, \orgaddress{\street{Street}, \city{City}, \postcode{610101}, \state{State}, \country{Country}}}


\abstract{
 Whole-body hemodynamics simulators, which model blood flow and pressure waveforms as functions of physiological parameters, are now essential tools for studying cardiovascular systems. However, solving the corresponding inverse problem of mapping observations (e.g., arterial pressure waveforms at specific locations in the arterial network) back to plausible physiological parameters remains challenging. Leveraging recent advances in simulation-based inference, we cast this problem as statistical inference by training an amortized neural posterior estimator on a newly built large dataset of cardiac simulations that we publicly release. To better align simulated data with real-world measurements, we incorporate stochastic elements modeling exogenous effects. The proposed framework can further integrate in-vivo data sources to refine its predictive capabilities on real-world data. In silico, we demonstrate that the proposed framework enables finely quantifying uncertainty associated with individual measurements, allowing trustworthy prediction of four biomarkers of clinical interest—namely Heart Rate, Cardiac Output, Systemic Vascular Resistance, and Left Ventricular Ejection Time—from arterial pressure waveforms and photoplethysmograms. Furthermore, we validate the framework in vivo, where our method accurately captures temporal trends in CO and SVR monitoring on the VitalDB dataset. Finally, the predictive error made by the model monotonically increases with the predicted uncertainty, thereby directly supporting the automatic rejection of unusable measurements.

 % incorporating stochastic noise modeling exogenous effects, which reduces the discrepancy between simulated and real-world measurements
 %We further propose an in-vivo model's refinement to further improve the estimation of cardiac parameters by reducing the bias in the predictions, whenever a calibration dataset is available.
}


\maketitle
\input{tex/introduction}

\input{tex/results}

\input{tex/results_invivo}

\input{tex/discussions}
\input{tex/method}

\input{tex/conclusion}

% \backmatter
\newpage
\bibliography{sn-bibliography}
\newpage

\bmhead{Supplementary information}
\input{tex/appendix}
% 



\end{document}

%% file: tex/introduction.tex
\section{Introduction}\label{sec1}
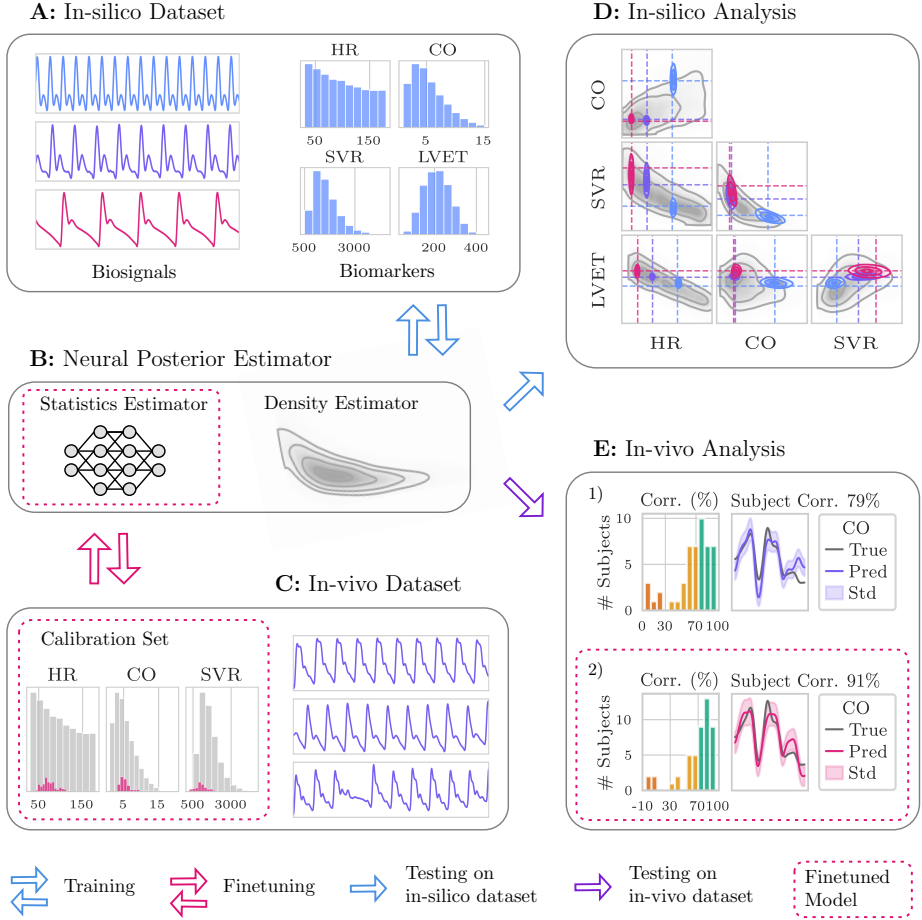
\begin{figure*}
  \input{figures/tikz/overview_figure}
  \caption{ Proposed framework for predicting cardiovascular parameters from cardiac measurements (\emph{biosignals}). Results are shown using arterial radial pressure waveforms (PWs), though PPGs can also be used.
  \textbf{A}: We generate a large-scale dataset of CV simulations using whole-body 1D hemodynamics simulators and appropriate noise models (see \autoref{fig:pre-processing} for more details). 
  \textbf{B}: A neural posterior estimator is trained on the in-silico dataset. It learns a surrogate of the posterior distribution of the parameters of interest given the PWs.
  \textbf{C}: The in-vivo dataset is comprised of PWs used to test the model on real-world distributions (gray histogram)and a small calibration set with labels used to finetune part of the model (pink histogram).
  \textbf{D}: The proposed framework is tested in-silico. The corner plot shows the learned posterior distributions of cardiac biomarkers for three colored waveforms from panel A, the prior distribution is shown in gray.
  \textbf{E}: The proposed framework is further validated in-vivo. The upper plot shows results on the model trained in-silico, while the lower plot shows results on the model further fine-tuned on the calibration set. For each plot, we show (left) the histogram of per-patient Spearman correlation between the measured and predicted
CO, and (right) the normalized predicted vs measured CO of a patient through time.
  }\label{fig:SBI_CV}
    % \vspace{-1.5em}
\end{figure*}

A comprehensive understanding of the human cardiovascular (CV) system is crucial to the prevention and management of cardiovascular diseases, which remain the leading cause of mortality worldwide \citep{10.1093/cvr/cvac013}. 
Over the past decades, there has been significant progress in CV modeling~\citep{stergiopulos1992computer, ribatti2009william}, evolving from manual calculations~\citep{altschule1938effects, patterson1914regulation} to advanced simulators~\citep{updegrove2017simvascular, melis2017gaussian, charlton2019modeling, alastruey2023arterial} that leverage modern scientific computing. Nowadays, CV simulators serve as an essential tool in the study of complex CV processes~\citep{PEGOLOTTI2024107676, kissas2023towards, popp2024development},  ranging from 3D models of cardiac function~\citep{baillargeon2014living} to full-scale simulations of hemodynamics across the entire arterial network~\citep{melis2017gaussian, charlton2019modeling, alastruey2023arterial}. CV simulators have been successfully integrated in various medical applications. For example, 
they have been used to enhance surgical techniques for congenital heart disease \citep{marsden2009evaluation} and to optimize cardiovascular stent design \citep{Gundert2012OptimizationOC}. 

Nevertheless, the full potential of CV simulators in precision medicine remains largely unexploited. 
In each of the previously mentioned applications, CV simulators take known physiological parameters as input and generate the corresponding biosignals (i.e., forward modeling). However, to fully integrate CV simulators in clinical settings, it is of critical importance to be able to infer a person's physiological state from observed biosignals (i.e., solving the inverse problem). For example, inverting a CV simulator could enable accurate estimation of critical biomarkers that would otherwise be difficult to measure, such as cardiac output (CO) and systemic vascular resistance (SVR)~\citep{Ross2016ImportanceOA,Lu2022AssociationOS}.  
Currently, the gold standard methods for measuring CO and SVR are invasive, while less invasive alternatives tend to have high error rates
\cite{Sanders2019AccuracyAP,Khnen2020SystemicVR}.

Inverting a CV simulator presents a new set of challenges~\citep{quick2001infinite, nolte2022inverse}.  We specifically focus on whole-body 1D hemodynamics simulators, which describe the arterial pulse wave propagation into different arterial segments given cardiac, arterial, and vascular bed properties. Designed for efficient forward simulation via numerical solvers, these models are not suited for inversion because they function as black-box stochastic simulators, preventing the application of likelihood-based or gradient-based inference methods. Additionally, given the complexity of the forward model, some physiological parameters might not be uniquely identifiable, and reflecting this uncertainty in estimated physiological state becomes crucial.  

To address these challenges, we leverage recent advances in simulation-based inference~\citep[SBI,][]{cranmer2020frontier, tejero2020sbi}, which uses machine learning to perform statistical inference over the parameters of black-box simulators without requiring an explicit likelihood function or gradient information.
As illustrated in \autoref{fig:SBI_CV}, we develop a novel and robust framework that combines in-vivo and in-silico data sources with SBI. This framework is used to validate whole-body 1D hemodynamics simulators~\citep{melis2017gaussian, charlton2019modeling} for predicting cardiac parameters from biosignals in the form of arterial pressure waveforms (APWs) or digital photoplethysmograms (PPGs).
Our results demonstrate that the proposed method enables population-level uncertainty analysis of cardiac properties of interest and enhances the performance of CO and SVR estimation from APWs and PPGs in-vivo. 

Our contributions are as follows: \textbf{(i)} we create a large-scale database of CV simulations, including radial APWs and digital PPGs with noise models, to bridge the gap between in-silico and in-vivo measurements using 1D whole-body hemodynamics simulators; \textbf{(ii)} we train a neural posterior estimator~\citep{lueckmann2017flexible} on these simulations, enabling efficient \textit{Bayesian} inference for inverting hemodynamics simulators; \textbf{(iii)} we validate the simulator and inference method on in-silico and in-vivo data, demonstrating accurate monitoring of Heart Rate~(HR), CO, and SVR from VitalDB~\citep{Lee2022VitalDBAH} measurements, while quantifying uncertainty for each measurement; and \textbf{(iv)} we introduce a hybrid learning strategy that enhances accuracy when a labeled calibration set of real-world data is available.

%% file: figures/tikz/overview_figure.tex
\resizebox{0.95\linewidth}{!}{

\tikzset{every picture/.style={line width=0.75pt}} %set default line width to 0.75pt        

\begin{tikzpicture}[x=0.75pt,y=0.75pt,yscale=-1,xscale=1]
%uncomment if require: \path (0,646); %set diagram left start at 0, and has height of 646

%Image [id:dp08952607590893369] 
\draw (524.5,392.5) node  {\includegraphics[width=169.5pt,height=90.75pt]{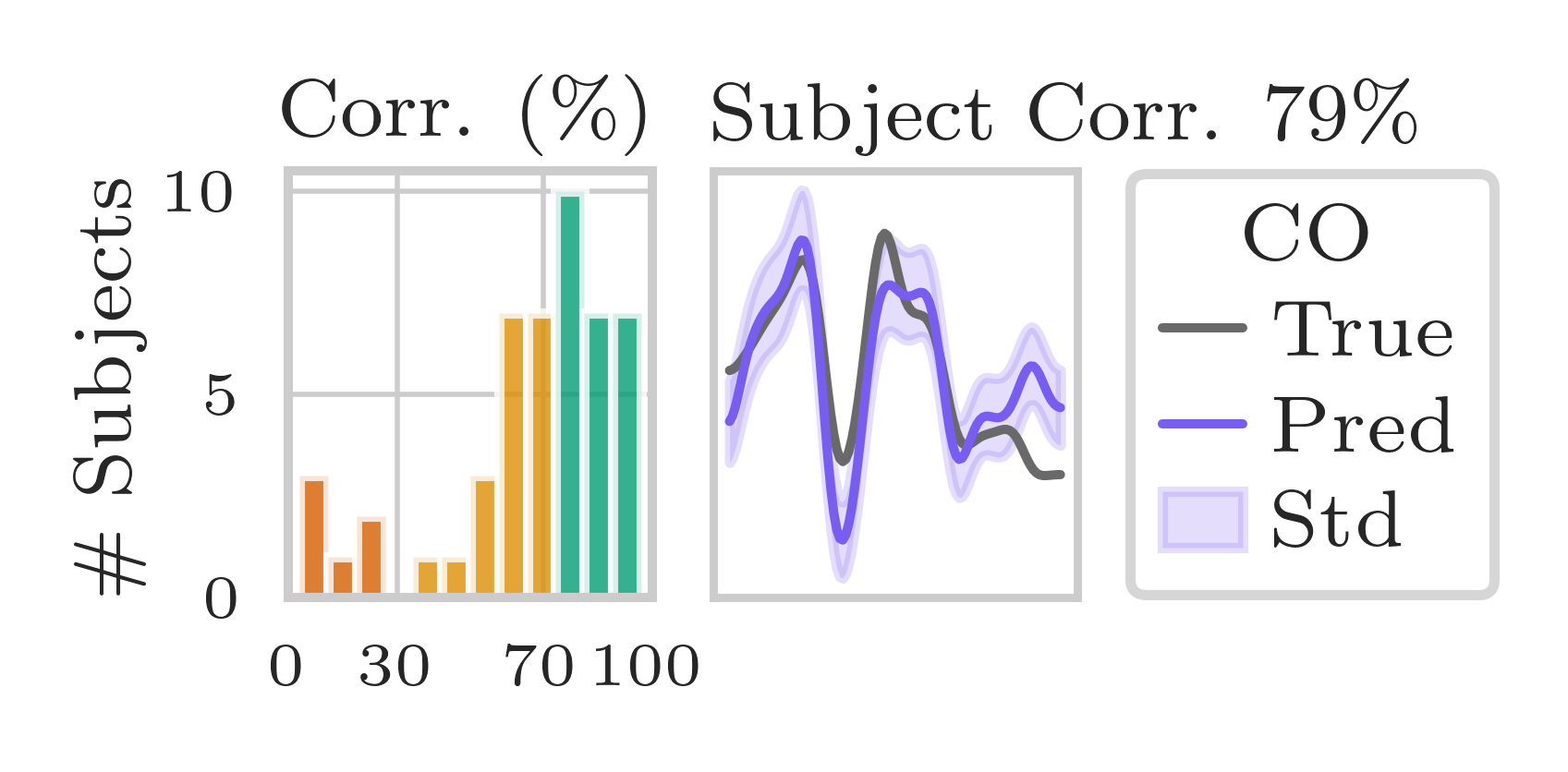}};
%Image [id:dp854798705617341] 
\draw (524.5,517.5) node  {\includegraphics[width=169.5pt,height=90.75pt]{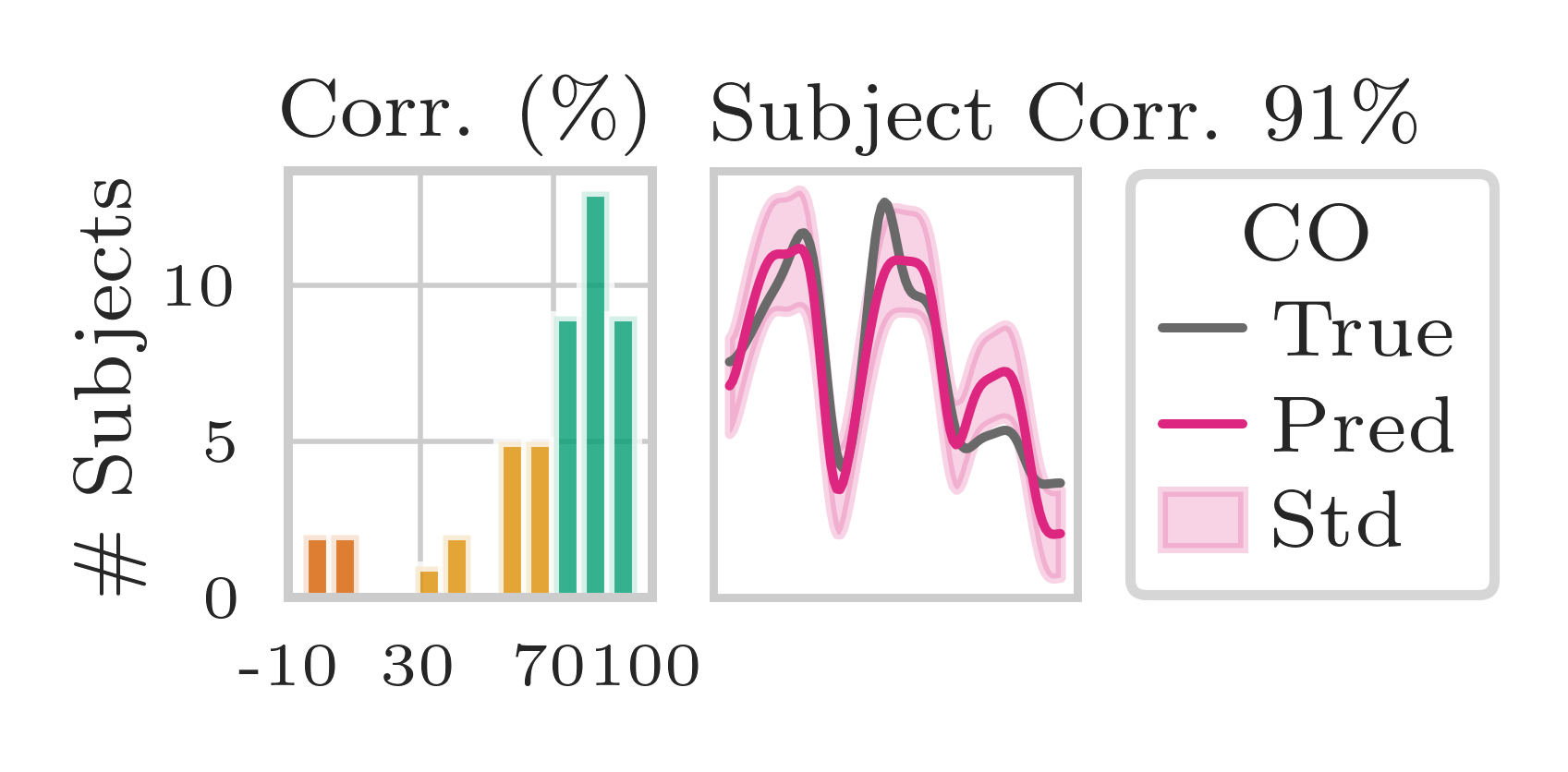}};
%Image [id:dp9625892161445074] 
\draw (107.6,515.46) node  {\includegraphics[width=135.6pt,height=90.07pt]{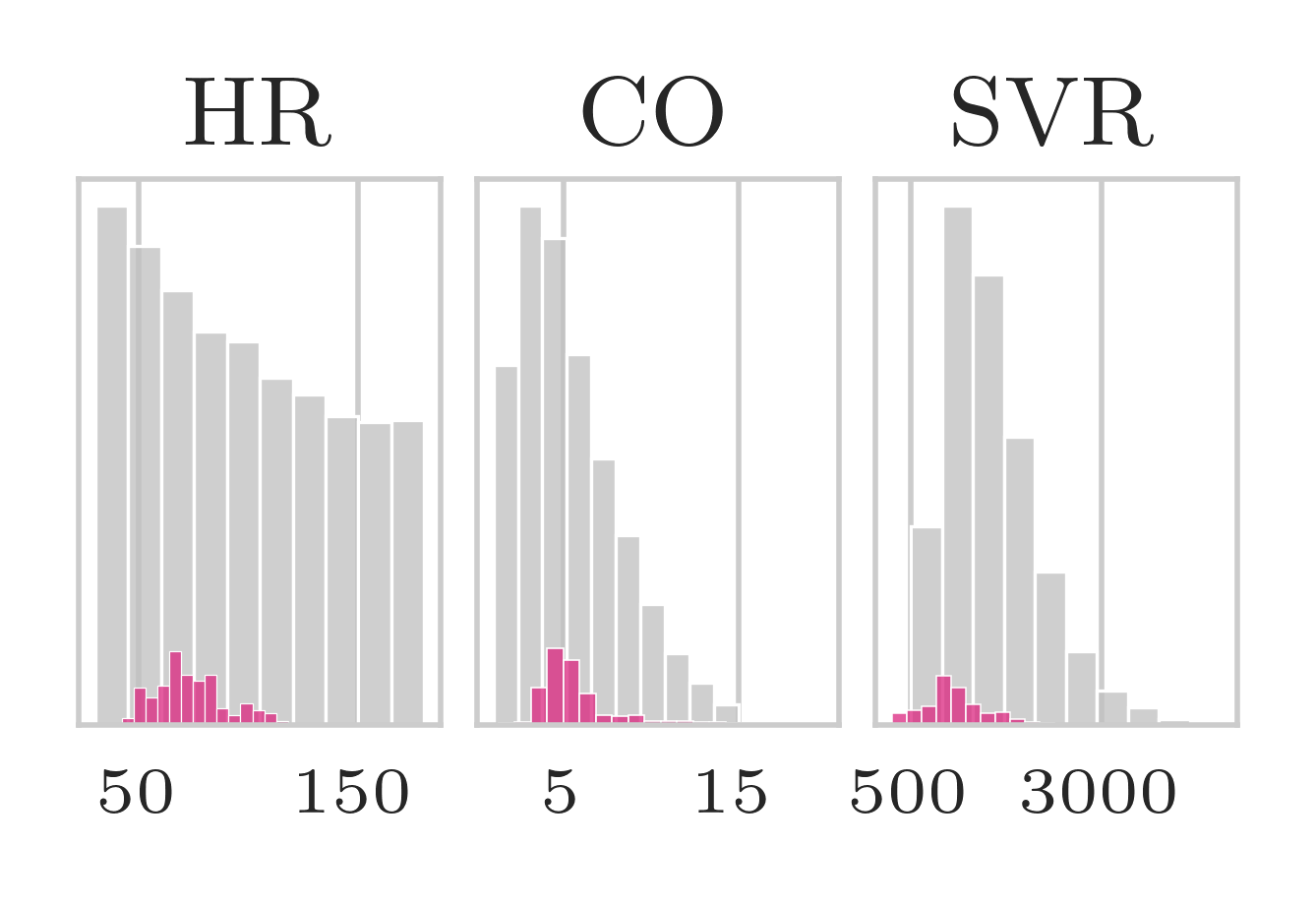}};
%Image [id:dp862218661509002] 
\draw (280,105) node  {\includegraphics[width=120pt,height=124.5pt]{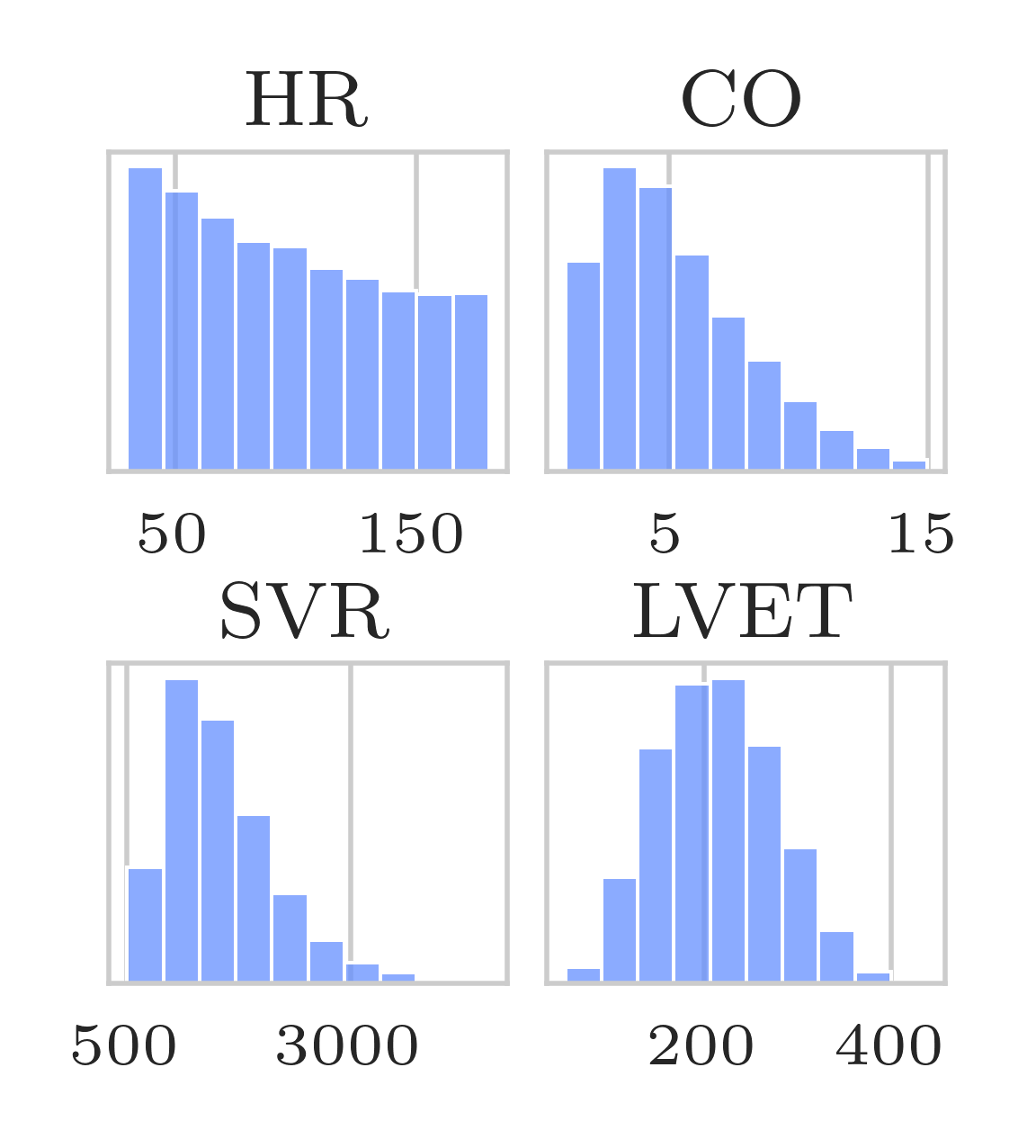}};
%Image [id:dp8923767019183769] 
\draw (524.75,141) node  {\includegraphics[width=182.62pt,height=180pt]{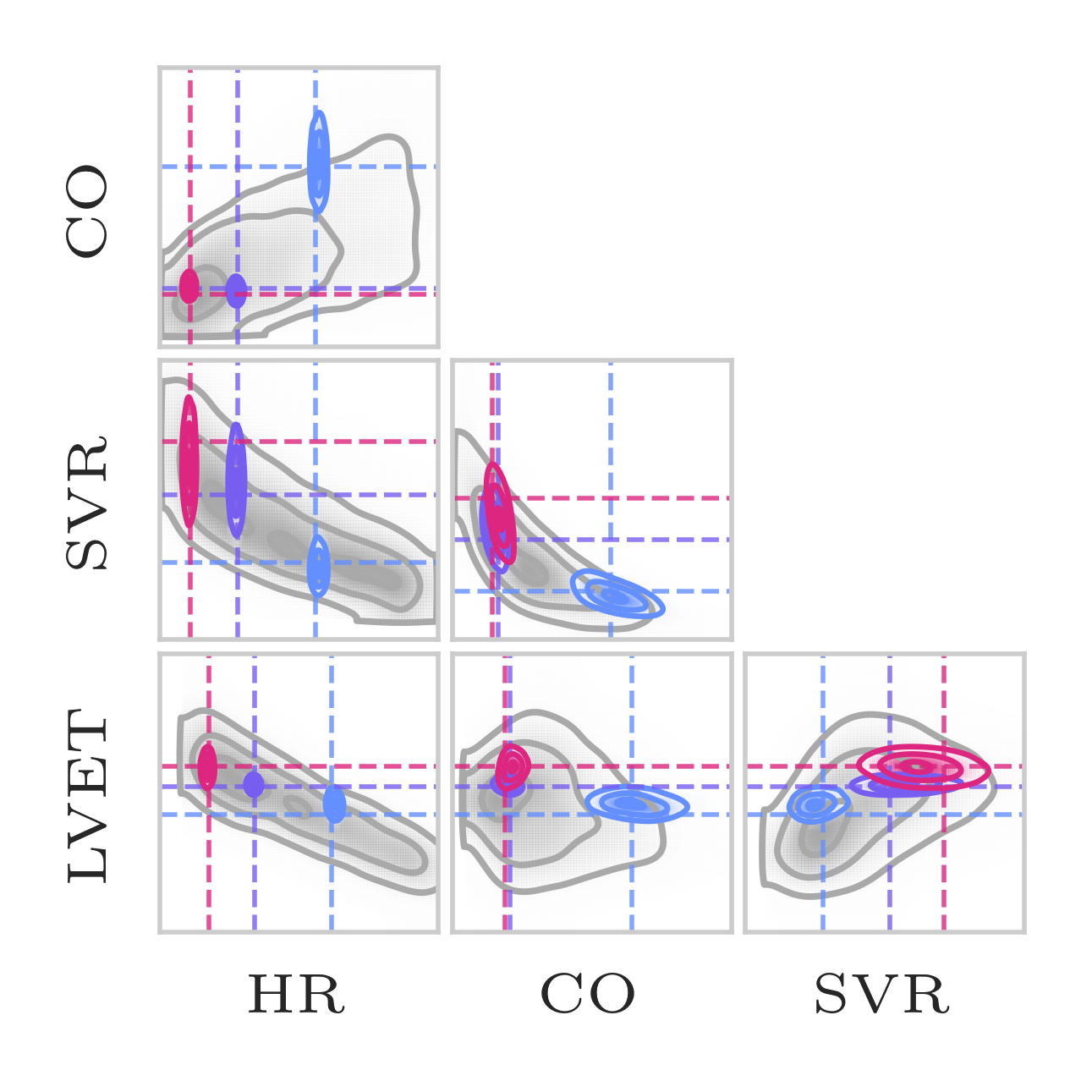}};
%Image [id:dp9478096076384831] 
\draw (268.75,306.5) node [rotate=-348.94] {\includegraphics[width=127.13pt,height=93.75pt]{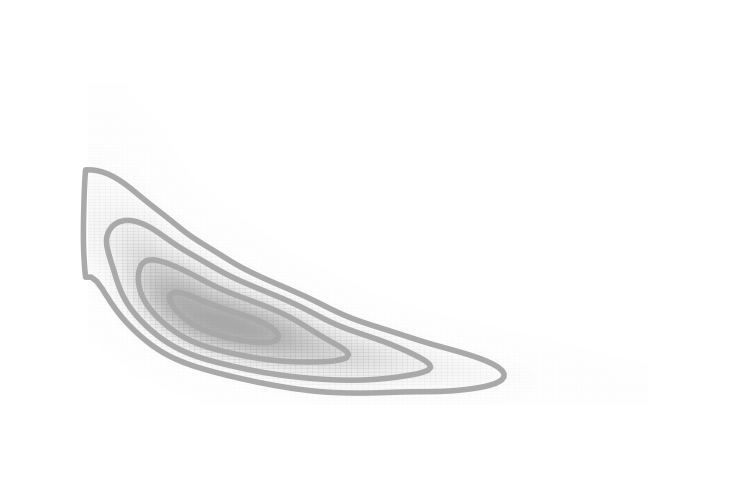}};
%Image [id:dp9611320147563344] 
\draw (104.25,106.25) node  {\includegraphics[width=118.88pt,height=118.88pt]{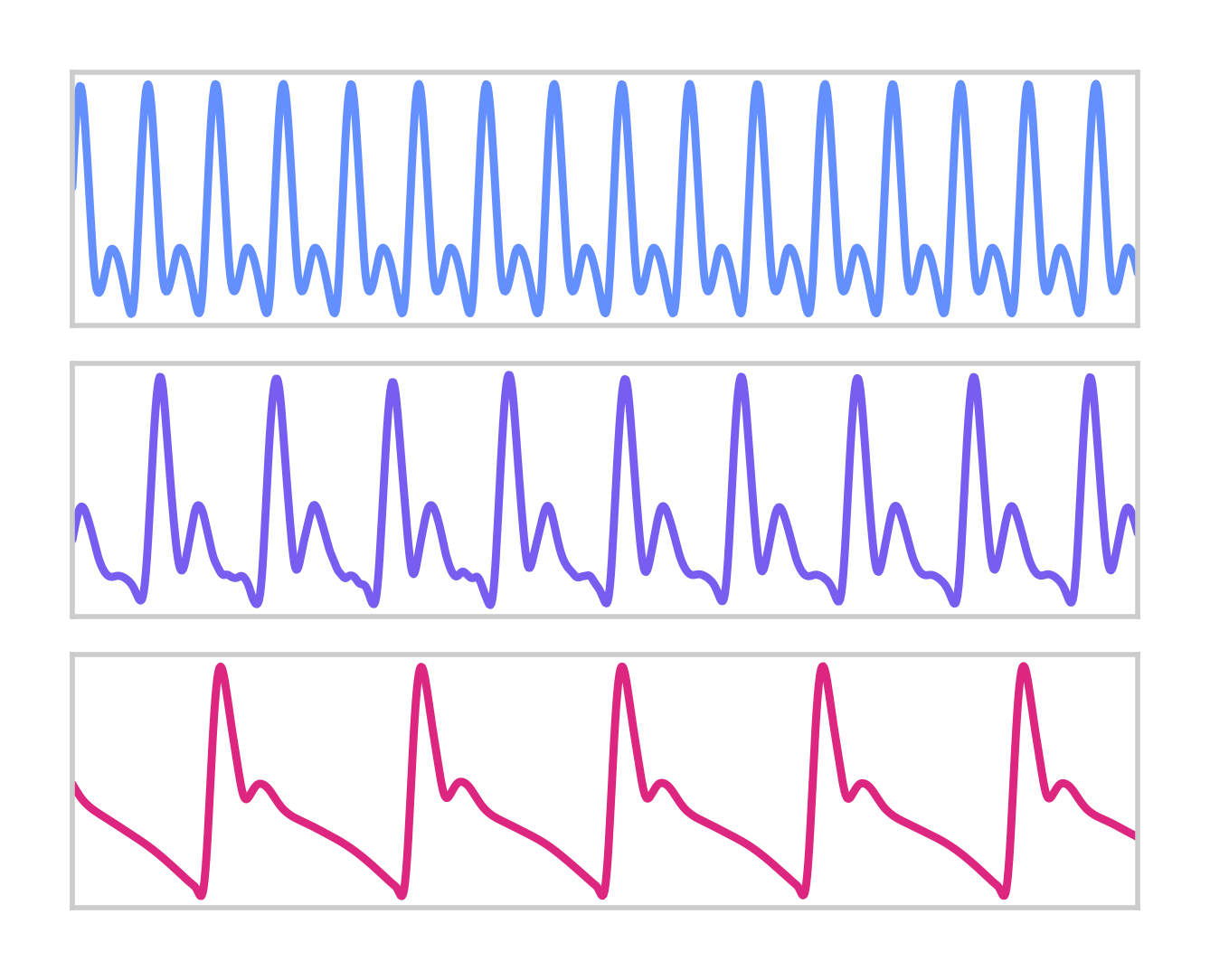}};
%Rounded Rect [id:dp24890785146129568] 
\draw  [color=black  ,draw opacity=.5 ][line width=.75]  (400,347.72) .. controls (400,337.7) and (408.12,329.58) .. (418.14,329.58) -- (625.64,329.58) .. controls (635.65,329.58) and (643.77,337.7) .. (643.77,347.72) -- (643.77,560.12) .. controls (643.77,570.14) and (635.65,578.26) .. (625.64,578.26) -- (418.14,578.26) .. controls (408.12,578.26) and (400,570.14) .. (400,560.12) -- cycle ;
%Rounded Rect [id:dp09368375289793007] 
\draw  [color={rgb, 255:red, 224; green, 16; blue, 114 }  ,draw opacity=1 ][dash pattern={on 1.69pt off 2.76pt}][line width=1.]  (407,460.5) .. controls (407,456.36) and (410.36,453) .. (414.5,453) -- (628.5,453) .. controls (632.64,453) and (636,456.36) .. (636,460.5) -- (636,564) .. controls (636,568.14) and (632.64,571.5) .. (628.5,571.5) -- (414.5,571.5) .. controls (410.36,571.5) and (407,568.14) .. (407,564) -- cycle ;
%Right Arrow [id:dp03383042702176087] 
\draw  [color={rgb, 255:red, 74; green, 144; blue, 226 }  ,draw opacity=1 ][line width=1.]  (291.5,247.93) -- (291.62,224.03) -- (283.56,223.99) -- (294.39,213.5) -- (305.12,224.09) -- (297.06,224.05) -- (296.95,247.96) -- cycle ;
%Right Arrow [id:dp32654837331109854] 
\draw  [color={rgb, 255:red, 74; green, 144; blue, 226 }  ,draw opacity=1 ][line width=1.]  (316.84,213.41) -- (317.27,237.31) -- (325.32,237.17) -- (314.73,247.89) -- (303.77,237.55) -- (311.82,237.41) -- (311.39,213.5) -- cycle ;

%Rounded Rect [id:dp2822000553759223] 
\draw  [color={rgb, 255:red, 224; green, 16; blue, 114 }  ,draw opacity=1 ][dash pattern={on 1.69pt off 2.76pt}][line width=1.]  (557.5,607.22) .. controls (557.5,603.48) and (560.53,600.46) .. (564.26,600.46) -- (631.74,600.46) .. controls (635.47,600.46) and (638.5,603.48) .. (638.5,607.22) -- (638.5,632.24) .. controls (638.5,635.97) and (635.47,639) .. (631.74,639) -- (564.26,639) .. controls (560.53,639) and (557.5,635.97) .. (557.5,632.24) -- cycle ;
%Rounded Rect [id:dp41535045625182565] 
\draw  [color=black,draw opacity=.5 ][line width=.75]  (400,45.59) .. controls (400,34.22) and (409.22,25) .. (420.59,25) -- (622.91,25) .. controls (634.28,25) and (643.5,34.22) .. (643.5,45.59) -- (643.5,236.41) .. controls (643.5,247.78) and (634.28,257) .. (622.91,257) -- (420.59,257) .. controls (409.22,257) and (400,247.78) .. (400,236.41) -- cycle ;
%Rounded Rect [id:dp7838788653814077] 
\draw  [color=black,draw opacity=.5 ][line width=.75] (14,40.89) .. controls (14,32.11) and (21.11,25) .. (29.89,25) -- (352.11,25) .. controls (360.89,25) and (368,32.11) .. (368,40.89) -- (368,188.11) .. controls (368,196.89) and (360.89,204) .. (352.11,204) -- (29.89,204) .. controls (21.11,204) and (14,196.89) .. (14,188.11) -- cycle ;
%Rounded Rect [id:dp16738470792746685] 
\draw  [color=black,draw opacity=.5 ][line width=.75]  (15.19,285) .. controls (15.19,275) and (23.3,266.89) .. (33.31,266.89) -- (315.68,266.89) .. controls (325.68,266.89) and (333.79,275) .. (333.79,285) -- (333.79,339.33) .. controls (333.79,349.34) and (325.68,357.45) .. (315.68,357.45) -- (33.31,357.45) .. controls (23.3,357.45) and (15.19,349.34) .. (15.19,339.33) -- cycle ;
%Image [id:dp34116809129449455] 
\draw (279.25,506.75) node  {\includegraphics[width=114.38pt,height=110.63pt]{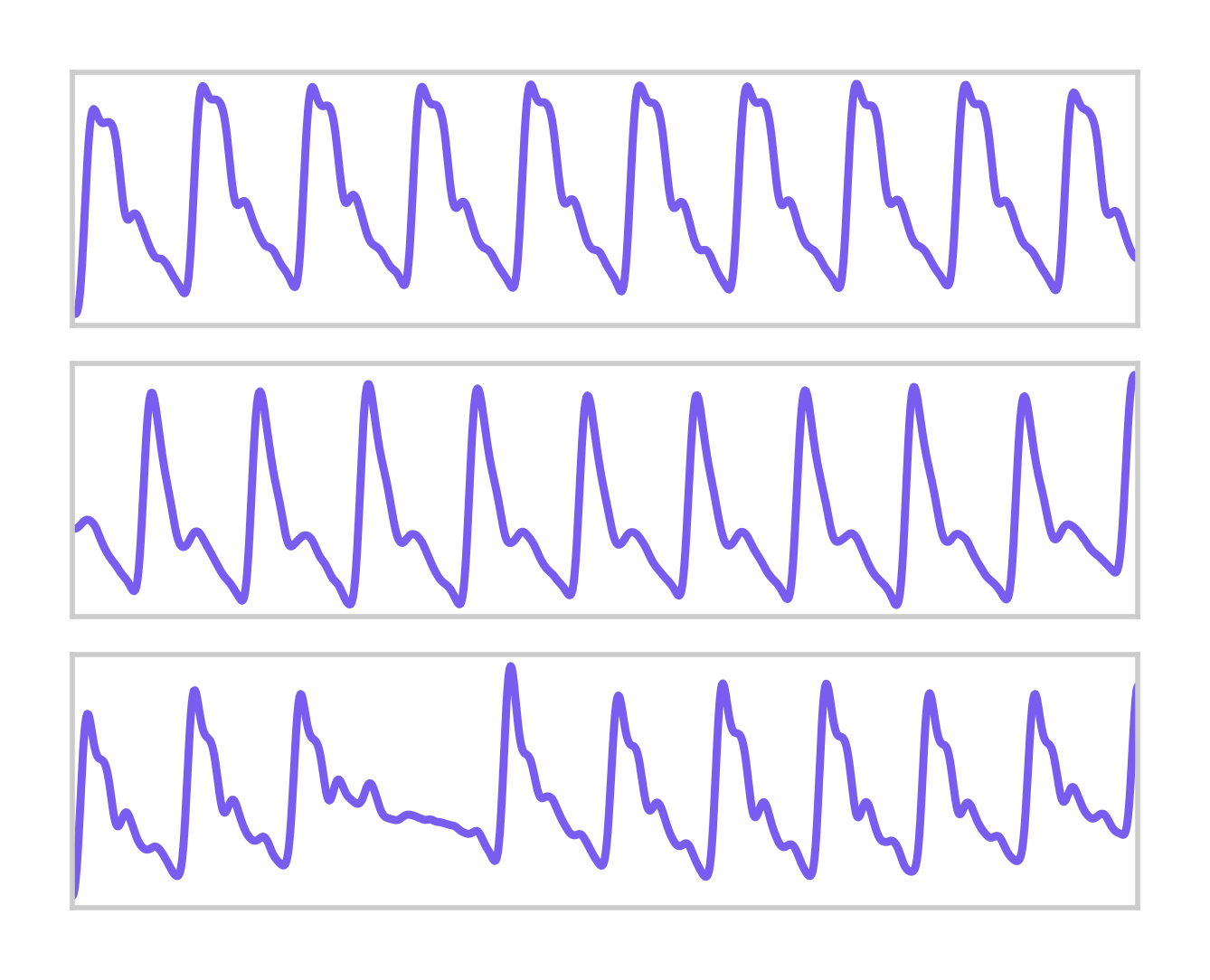}};
%Rounded Rect [id:dp0812941468507814] 
\draw  [color=black,draw opacity=.5 ][line width=.75]  (14,440.56) .. controls (14,431.42) and (21.42,424) .. (30.56,424) -- (343.44,424) .. controls (352.58,424) and (360,431.42) .. (360,440.56) -- (360,561.69) .. controls (360,570.84) and (352.58,578.26) .. (343.44,578.26) -- (30.56,578.26) .. controls (21.42,578.26) and (14,570.84) .. (14,561.69) -- cycle ;
%Rounded Rect [id:dp9047221483389777] 
\draw  [color={rgb, 255:red, 224; green, 16; blue, 114 }  ,draw opacity=1 ][dash pattern={on 1.69pt off 2.76pt}][line width=1.]  (25.31,277.43) .. controls (25.31,274.64) and (27.56,272.39) .. (30.34,272.39) -- (155.46,272.39) .. controls (158.24,272.39) and (160.5,274.64) .. (160.5,277.43) -- (160.5,346.96) .. controls (160.5,349.74) and (158.24,352) .. (155.46,352) -- (30.34,352) .. controls (27.56,352) and (25.31,349.74) .. (25.31,346.96) -- cycle ;
%Rounded Rect [id:dp5950828721692966] 
\draw  [color={rgb, 255:red, 224; green, 16; blue, 114 }  ,draw opacity=1 ][dash pattern={on 1.69pt off 2.76pt}][line width=1.]  (22.31,441.16) .. controls (22.31,436.31) and (26.23,432.39) .. (31.08,432.39) -- (185.73,432.39) .. controls (190.57,432.39) and (194.5,436.31) .. (194.5,441.16) -- (194.5,562.23) .. controls (194.5,567.07) and (190.57,571) .. (185.73,571) -- (31.08,571) .. controls (26.23,571) and (22.31,567.07) .. (22.31,562.23) -- cycle ;
%Right Arrow [id:dp9062395720600102] 
\draw  [color={rgb, 255:red, 74; green, 144; blue, 226 }  ,draw opacity=1 ][line width=1.]  (357.19,282.74) -- (374.31,266.06) -- (368.69,260.29) -- (383.76,260.66) -- (383.73,275.73) -- (378.11,269.96) -- (360.99,286.64) -- cycle ;
%Right Arrow [id:dp31917530814452677] 
\draw  [color={rgb, 255:red, 123; green, 29; blue, 209 }  ,draw opacity=1 ][line width=1.]  (361.14,333.69) -- (378.07,350.58) -- (383.75,344.87) -- (383.6,359.94) -- (368.53,360.13) -- (374.22,354.43) -- (357.29,337.55) -- cycle ;
%Right Arrow [id:dp8863550397241187] 
\draw  [color={rgb, 255:red, 224; green, 16; blue, 114 }  ,draw opacity=1 ][line width=1.]  (71.5,406.93) -- (71.62,383.03) -- (63.56,382.99) -- (74.39,372.5) -- (85.12,383.09) -- (77.06,383.05) -- (76.95,406.96) -- cycle ;
%Right Arrow [id:dp2647753494626771] 
\draw  [color={rgb, 255:red, 224; green, 16; blue, 114 }  ,draw opacity=1 ][line width=1.]  (96.84,372.41) -- (97.27,396.31) -- (105.32,396.17) -- (94.73,406.89) -- (83.77,396.55) -- (91.82,396.41) -- (91.39,372.5) -- cycle ;

%Right Arrow [id:dp7875153121025099] 
\draw  [color={rgb, 255:red, 224; green, 16; blue, 114 }  ,draw opacity=1 ][line width=1.]  (151.53,631.89) -- (134.63,631.67) -- (134.55,637.87) -- (127.21,629.47) -- (134.76,621.28) -- (134.68,627.47) -- (151.58,627.7) -- cycle ;
%Right Arrow [id:dp7452478469582554] 
\draw  [color={rgb, 255:red, 224; green, 16; blue, 114 }  ,draw opacity=1 ][line width=1.]  (127.28,612.2) -- (144.18,612) -- (144.13,605.8) -- (151.64,614.01) -- (144.27,622.39) -- (144.22,616.19) -- (127.32,616.39) -- cycle ;

%Right Arrow [id:dp7667350516892214] 
\draw  [color={rgb, 255:red, 74; green, 144; blue, 226 }  ,draw opacity=1 ][line width=1.]  (41.53,630.89) -- (24.63,630.67) -- (24.55,636.87) -- (17.21,628.47) -- (24.76,620.28) -- (24.68,626.47) -- (41.58,626.7) -- cycle ;
%Right Arrow [id:dp7984998226243673] 
\draw  [color={rgb, 255:red, 74; green, 144; blue, 226 }  ,draw opacity=1 ][line width=1.]  (17.28,611.2) -- (34.18,611) -- (34.13,604.8) -- (41.64,613.01) -- (34.27,621.39) -- (34.22,615.19) -- (17.32,615.39) -- cycle ;

%Right Arrow [id:dp03730531101737067] 
\draw  [color={rgb, 255:red, 74; green, 144; blue, 226 }  ,draw opacity=1 ][line width=1.]  (251.28,617.2) -- (268.18,617) -- (268.13,610.8) -- (275.64,619.01) -- (268.27,627.39) -- (268.22,621.19) -- (251.32,621.39) -- cycle ;
%Right Arrow [id:dp08740017344810314] 
\draw  [color={rgb, 255:red, 123; green, 29; blue, 209 }  ,draw opacity=1 ][line width=1.]  (406.28,616.2) -- (423.18,616) -- (423.13,609.8) -- (430.64,618.01) -- (423.27,626.39) -- (423.22,620.19) -- (406.32,620.39) -- cycle ;
%Shape: Ellipse [id:dp5879867357131159] 
\draw  [fill={rgb, 255:red, 0; green, 0; blue, 0 }  ,fill opacity=0.12 ][line width=0.75]  (54,314.89) .. controls (54,312.51) and (55.95,310.59) .. (58.35,310.59) .. controls (60.76,310.59) and (62.71,312.51) .. (62.71,314.89) .. controls (62.71,317.26) and (60.76,319.19) .. (58.35,319.19) .. controls (55.95,319.19) and (54,317.26) .. (54,314.89) -- cycle ;
%Shape: Ellipse [id:dp45451963051936517] 
\draw  [fill={rgb, 255:red, 0; green, 0; blue, 0 }  ,fill opacity=0.12 ][line width=0.75]  (54,328.12) .. controls (54,325.74) and (55.95,323.82) .. (58.35,323.82) .. controls (60.76,323.82) and (62.71,325.74) .. (62.71,328.12) .. controls (62.71,330.49) and (60.76,332.42) .. (58.35,332.42) .. controls (55.95,332.42) and (54,330.49) .. (54,328.12) -- cycle ;
%Shape: Ellipse [id:dp6616478608184652] 
\draw  [fill={rgb, 255:red, 0; green, 0; blue, 0 }  ,fill opacity=0.12 ][line width=0.75]  (74.1,341.35) .. controls (74.1,338.98) and (76.05,337.05) .. (78.45,337.05) .. controls (80.86,337.05) and (82.81,338.98) .. (82.81,341.35) .. controls (82.81,343.73) and (80.86,345.65) .. (78.45,345.65) .. controls (76.05,345.65) and (74.1,343.73) .. (74.1,341.35) -- cycle ;
%Shape: Ellipse [id:dp8833305053493802] 
\draw  [fill={rgb, 255:red, 0; green, 0; blue, 0 }  ,fill opacity=0.12 ][line width=0.75]  (74.1,328.12) .. controls (74.1,325.74) and (76.05,323.82) .. (78.45,323.82) .. controls (80.86,323.82) and (82.81,325.74) .. (82.81,328.12) .. controls (82.81,330.49) and (80.86,332.42) .. (78.45,332.42) .. controls (76.05,332.42) and (74.1,330.49) .. (74.1,328.12) -- cycle ;
%Shape: Ellipse [id:dp9240407507344568] 
\draw  [fill={rgb, 255:red, 0; green, 0; blue, 0 }  ,fill opacity=0.12 ][line width=0.75]  (74.1,314.89) .. controls (74.1,312.51) and (76.05,310.59) .. (78.45,310.59) .. controls (80.86,310.59) and (82.81,312.51) .. (82.81,314.89) .. controls (82.81,317.26) and (80.86,319.19) .. (78.45,319.19) .. controls (76.05,319.19) and (74.1,317.26) .. (74.1,314.89) -- cycle ;
%Shape: Ellipse [id:dp7950656450459244] 
\draw  [fill={rgb, 255:red, 0; green, 0; blue, 0 }  ,fill opacity=0.12 ][line width=0.75]  (74.1,301.65) .. controls (74.1,299.28) and (76.05,297.35) .. (78.45,297.35) .. controls (80.86,297.35) and (82.81,299.28) .. (82.81,301.65) .. controls (82.81,304.03) and (80.86,305.96) .. (78.45,305.96) .. controls (76.05,305.96) and (74.1,304.03) .. (74.1,301.65) -- cycle ;
%Shape: Ellipse [id:dp44074368390736207] 
\draw  [fill={rgb, 255:red, 0; green, 0; blue, 0 }  ,fill opacity=0.12 ][line width=0.75]  (114.29,314.89) .. controls (114.29,312.51) and (116.24,310.59) .. (118.65,310.59) .. controls (121.05,310.59) and (123,312.51) .. (123,314.89) .. controls (123,317.26) and (121.05,319.19) .. (118.65,319.19) .. controls (116.24,319.19) and (114.29,317.26) .. (114.29,314.89) -- cycle ;
%Shape: Ellipse [id:dp6121159873216382] 
\draw  [fill={rgb, 255:red, 0; green, 0; blue, 0 }  ,fill opacity=0.12 ][line width=0.75]  (94.19,341.35) .. controls (94.19,338.98) and (96.14,337.05) .. (98.55,337.05) .. controls (100.95,337.05) and (102.9,338.98) .. (102.9,341.35) .. controls (102.9,343.73) and (100.95,345.65) .. (98.55,345.65) .. controls (96.14,345.65) and (94.19,343.73) .. (94.19,341.35) -- cycle ;
%Shape: Ellipse [id:dp5425330488200347] 
\draw  [fill={rgb, 255:red, 0; green, 0; blue, 0 }  ,fill opacity=0.12 ][line width=0.75]  (94.19,328.12) .. controls (94.19,325.74) and (96.14,323.82) .. (98.55,323.82) .. controls (100.95,323.82) and (102.9,325.74) .. (102.9,328.12) .. controls (102.9,330.49) and (100.95,332.42) .. (98.55,332.42) .. controls (96.14,332.42) and (94.19,330.49) .. (94.19,328.12) -- cycle ;
%Shape: Ellipse [id:dp9504298832681006] 
\draw  [fill={rgb, 255:red, 0; green, 0; blue, 0 }  ,fill opacity=0.12 ][line width=0.75]  (94.19,314.89) .. controls (94.19,312.51) and (96.14,310.59) .. (98.55,310.59) .. controls (100.95,310.59) and (102.9,312.51) .. (102.9,314.89) .. controls (102.9,317.26) and (100.95,319.19) .. (98.55,319.19) .. controls (96.14,319.19) and (94.19,317.26) .. (94.19,314.89) -- cycle ;
%Shape: Ellipse [id:dp340165739920846] 
\draw  [fill={rgb, 255:red, 0; green, 0; blue, 0 }  ,fill opacity=0.12 ][line width=0.75]  (94.19,301.65) .. controls (94.19,299.28) and (96.14,297.35) .. (98.55,297.35) .. controls (100.95,297.35) and (102.9,299.28) .. (102.9,301.65) .. controls (102.9,304.03) and (100.95,305.96) .. (98.55,305.96) .. controls (96.14,305.96) and (94.19,304.03) .. (94.19,301.65) -- cycle ;
%Shape: Ellipse [id:dp9765490896405781] 
\draw  [fill={rgb, 255:red, 0; green, 0; blue, 0 }  ,fill opacity=0.12 ][line width=0.75]  (114.29,328.12) .. controls (114.29,325.74) and (116.24,323.82) .. (118.65,323.82) .. controls (121.05,323.82) and (123,325.74) .. (123,328.12) .. controls (123,330.49) and (121.05,332.42) .. (118.65,332.42) .. controls (116.24,332.42) and (114.29,330.49) .. (114.29,328.12) -- cycle ;
%Straight Lines [id:da43730453898616983] 
\draw [fill={rgb, 255:red, 0; green, 0; blue, 0 }  ,fill opacity=0.12 ][line width=0.75]    (62.71,314.89) -- (74.1,301.65) ;
%Straight Lines [id:da06719521861703615] 
\draw [fill={rgb, 255:red, 0; green, 0; blue, 0 }  ,fill opacity=0.12 ][line width=0.75]    (62.71,314.89) -- (74.1,314.89) ;
%Straight Lines [id:da7483061710116851] 
\draw [fill={rgb, 255:red, 0; green, 0; blue, 0 }  ,fill opacity=0.12 ][line width=0.75]    (62.71,328.12) -- (74.1,314.89) ;
%Straight Lines [id:da5606012861021814] 
\draw [fill={rgb, 255:red, 0; green, 0; blue, 0 }  ,fill opacity=0.12 ][line width=0.75]    (82.81,328.12) -- (94.19,314.89) ;
%Straight Lines [id:da7813243219986002] 
\draw [fill={rgb, 255:red, 0; green, 0; blue, 0 }  ,fill opacity=0.12 ][line width=0.75]    (82.81,301.65) -- (94.19,301.65) ;
%Straight Lines [id:da04874173800863424] 
\draw [fill={rgb, 255:red, 0; green, 0; blue, 0 }  ,fill opacity=0.12 ][line width=0.75]    (62.71,328.12) -- (74.1,341.35) ;
%Straight Lines [id:da4064970154410721] 
\draw [fill={rgb, 255:red, 0; green, 0; blue, 0 }  ,fill opacity=0.12 ][line width=0.75]    (62.71,314.89) -- (74.1,328.12) ;
%Straight Lines [id:da9622400739924859] 
\draw [fill={rgb, 255:red, 0; green, 0; blue, 0 }  ,fill opacity=0.12 ][line width=0.75]    (62.71,328.12) -- (74.1,328.12) ;
%Straight Lines [id:da9526474742456149] 
\draw [fill={rgb, 255:red, 0; green, 0; blue, 0 }  ,fill opacity=0.12 ][line width=0.75]    (102.9,314.89) -- (114.29,328.12) ;
%Straight Lines [id:da6470372583302061] 
\draw [fill={rgb, 255:red, 0; green, 0; blue, 0 }  ,fill opacity=0.12 ][line width=0.75]    (82.81,328.12) -- (94.19,328.12) ;
%Straight Lines [id:da2845265830116418] 
\draw [fill={rgb, 255:red, 0; green, 0; blue, 0 }  ,fill opacity=0.12 ][line width=0.75]    (82.81,314.89) -- (94.19,328.12) ;
%Straight Lines [id:da2753710889805957] 
\draw [fill={rgb, 255:red, 0; green, 0; blue, 0 }  ,fill opacity=0.12 ][line width=0.75]    (82.81,314.89) -- (94.19,314.89) ;
%Straight Lines [id:da9652567757441806] 
\draw [fill={rgb, 255:red, 0; green, 0; blue, 0 }  ,fill opacity=0.12 ][line width=0.75]    (82.81,301.65) -- (94.19,301.65) ;
%Straight Lines [id:da11560053252141367] 
\draw [fill={rgb, 255:red, 0; green, 0; blue, 0 }  ,fill opacity=0.12 ][line width=0.75]    (82.81,314.89) -- (94.19,301.65) ;
%Straight Lines [id:da0064055155574404] 
\draw [fill={rgb, 255:red, 0; green, 0; blue, 0 }  ,fill opacity=0.12 ][line width=0.75]    (82.81,301.65) -- (94.19,314.89) ;
%Straight Lines [id:da33726002289801704] 
\draw [fill={rgb, 255:red, 0; green, 0; blue, 0 }  ,fill opacity=0.12 ][line width=0.75]    (82.81,328.12) -- (94.19,341.35) ;
%Straight Lines [id:da4414625429007696] 
\draw [fill={rgb, 255:red, 0; green, 0; blue, 0 }  ,fill opacity=0.12 ][line width=0.75]    (82.81,341.35) -- (94.19,341.35) ;
%Straight Lines [id:da23397771858693805] 
\draw [fill={rgb, 255:red, 0; green, 0; blue, 0 }  ,fill opacity=0.12 ][line width=0.75]    (102.9,314.89) -- (114.29,314.89) ;
%Straight Lines [id:da15972421911683032] 
\draw [fill={rgb, 255:red, 0; green, 0; blue, 0 }  ,fill opacity=0.12 ][line width=0.75]    (102.9,328.12) -- (114.29,328.12) ;
%Straight Lines [id:da683016141474174] 
\draw [fill={rgb, 255:red, 0; green, 0; blue, 0 }  ,fill opacity=0.12 ][line width=0.75]    (82.81,341.35) -- (94.19,328.12) ;
%Straight Lines [id:da7685666752188972] 
\draw [fill={rgb, 255:red, 0; green, 0; blue, 0 }  ,fill opacity=0.12 ][line width=0.75]    (102.9,328.12) -- (114.29,314.89) ;
%Straight Lines [id:da19163685653571272] 
\draw [fill={rgb, 255:red, 0; green, 0; blue, 0 }  ,fill opacity=0.12 ][line width=0.75]    (102.9,341.35) -- (114.29,328.12) ;
%Straight Lines [id:da3322219514344106] 
\draw [fill={rgb, 255:red, 0; green, 0; blue, 0 }  ,fill opacity=0.12 ][line width=0.75]    (102.9,301.65) -- (114.29,314.89) ;

% Text Node
\draw (28,2) node [anchor=north west][inner sep=0.75pt]  [font=\large] [align=left] {\textbf{A:} In-silico Dataset};
% Text Node
\draw (417,2) node [anchor=north west][inner sep=0.75pt]  [font=\large] [align=left] {\textbf{D:} In-silico Analysis};
% Text Node
\draw (35.77,439) node [anchor=north west][inner sep=0.75pt]   [align=left] {Calibration Set};
% Text Node
\draw (197.46,400) node [anchor=north west][inner sep=0.75pt]  [font=\large] [align=left] {\textbf{C:} In-vivo Dataset};
% Text Node
\draw (161.64,611) node [anchor=north west][inner sep=0.75pt]   [align=left] {Finetuning};
% Text Node
\draw (28,243) node [anchor=north west][inner sep=0.75pt]  [font=\large] [align=left] {\textbf{B:} Neural Posterior Estimator};
% Text Node
\draw (35.31,275.89) node [anchor=north west][inner sep=0.75pt]  [font=\normalsize] [align=left] {Statistics Estimator};
% Text Node
\draw (190,275.27) node [anchor=north west][inner sep=0.75pt]  [font=\normalsize] [align=left] {Density Estimator};
% Text Node
\draw (417,306) node [anchor=north west][inner sep=0.75pt]  [font=\large] [align=left] {\textbf{E:} In-vivo Analysis};
% Text Node
\draw (413.5,338) node [anchor=north west][inner sep=0.75pt]  [font=\small] [align=left] {1) };
% Text Node
\draw (413.5,460) node [anchor=north west][inner sep=0.75pt]  [font=\small] [align=left] {2) };
% Text Node
\draw (51.64,611) node [anchor=north west][inner sep=0.75pt]   [align=left] {Training};
% Text Node
\draw (287.64,602) node [anchor=north west][inner sep=0.75pt]   [align=left] {Testing on \\in-silico dataset};
% Text Node
\draw (441.64,602) node [anchor=north west][inner sep=0.75pt]   [align=left] {Testing on \\in-vivo dataset};
% Text Node
\draw (563.18,604.46) node [anchor=north west][inner sep=0.75pt]   [align=left] {Finetuned \\Model};
% Text Node
\draw (242,183) node [anchor=north west][inner sep=0.75pt]   [align=left] {Biomarkers};
% Text Node
\draw (71,184) node [anchor=north west][inner sep=0.75pt]   [align=left] {Biosignals};

\end{tikzpicture}
}

%% file: tex/results.tex
\section{Results}\label{sec:results}
The primary objective of this study is to demonstrate that 1D hemodynamics simulations can be used to estimate cardiac parameter (i.e., \emph{biomarkers}) from standard-of-care measurements (i.e., \emph{biosignals}). 
% In the rest of the paper, we coin the terms \emph{biomarkers} and \emph{biosignals} to denote these parameters and the measurements used to infer them, respectively. 
While our methodology could apply to any biomarkers or biosignals modeled by the considered simulator, this study focuses on two standard-of-care biosignals: the arterial pressure waveforms (APWs) at the radial artery and the digital photoplethysmograms (PPGs). For biomarkers, we consider heart rate~\citep[HR,][]{kannel1987heart}, cardiac output~\citep[CO,][]{Sanders2019AccuracyAP}, systemic vascular resistance~\citep[SVR,][]{cotter2003role}, and left ventricular ejection time~\citep[LVET,][]{alhakak2021significance}. While, HR mostly serves as a baseline biomarker for designing the noise model and validating the proposed inference method, the three other biomarkers are chosen because of their relevance to assessing CV health~\citep{kannel1987heart, alhakak2021significance, patel2005cardiovascular, sutton2005elevated, cotter2003role}. 

In \autoref{fig:SBI_CV}, we provide an overview of the framework and achieved results. We first generate a large-scale dataset of approximately $80\text{,}000$ virtual subjects with uniform coverage of cardiovascular parameters, as detailed in \autoref{sec:simulations}, from which $8$-second segments of corresponding APWs and PPGs are generated. 
% After appropriate filtering, we use the whole-body 1D cardiovascular simulator to generate biosignals that are commonly collected in intensive care units (ICUs): the APW at the radial artery and the PPGs at the digital and the radial arteries. We couple them with an appropriate stochastic measurement model that randomly perturbs parts of the waveforms with various level of noise. 
 % For an in-depth description of the in-silico dataset generation, we refer to \autoref{sec:simulations}. 
 We then train a Neural Posterior Estimation (NPE), a Bayesian SBI method, that solves the inverse problem of mapping each biosignal back to the corresponding posterior distribution of biomarkers. 
 Herein, we validate the proposed framework both in-silico and in-vivo. 
In particular, in \autoref{sec:in-silico}, we assess the robustness of the method to noisy settings and the identifiability of the selected biomarkers for each considered biosignal in-silico. In \autoref{sec:in-vivo}, we perform in-vivo experiments to further validate NPE trained in-silico, on real-world measurements. Finally, we demonstrate that combining in-silico data with a small calibration set of in-vivo data further enhances the predictive performance and outperforms purely data-driven or physics-based approaches.

We split the in-silico dataset into train ($70\%$), validation ($10\%$), and test set ($20\%$) at random. All results reported, unless stated otherwise, are computed by training the NPE on the synthetic train set and selecting the best model based on the validation set likelihood, and tested on the in-silico or in-vivo test-set, where error bars report one standard deviation over five training instances. 

\subsection{In-silico analysis}\label{sec:in-silico}
In this section, we investigate the identifiability of cardiac biomarkers under various noise conditions, reducing the gap between in-vivo and in-silico measurements. 
To assess the noise level, we use the signal-to-noise ratio~(SNR), which is a measure of the strength of the signal relative to the noise. As metrics, we use the mean absolute error~(MAE) between the mean of the predicted biomarkers' posterior distribution and their ground truth values, the average coverage AUC~(ACAUC) of the credible intervals extracted from the estimated posteriors, and the average size of the credible intervals~(SCI). The ACAUC quantifies how accurately the predicted posterior probabilities align with the true underlying probabilities: it equals 0 under perfect calibration and
is bounded by 0.5. The SCI is the expected size of the predicted biomarker credible region at a certain level $\alpha$. These two metrics are further described and motivated in \autoref{sec:metrics}.

\paragraph{Designing the measurement model.}
\begin{figure*}
    \centering
    \includegraphics[width=1.\textwidth]{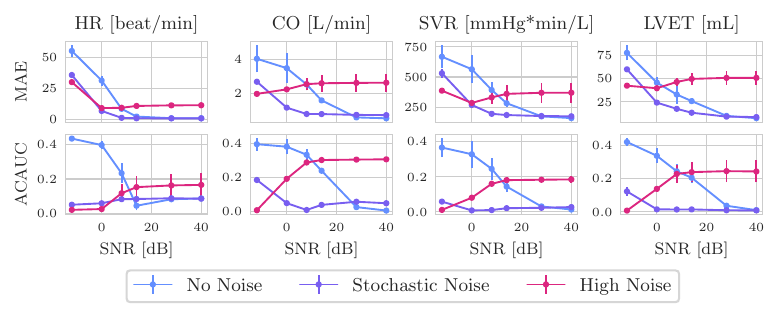}
    % {figures/combined_metrics_vs_SNR.pdf}
    \caption{MAE and ACAUC of the learned posterior distribution at different SNR. Results are reported for the NPE algorithm trained on synthetic APWs with (a) no noise model, (b) stochastic noise, and (c) fixed high noise. 
    }
    \label{fig:synthetic_SNR_metrics}
    \vspace{-1.5em}
\end{figure*}
To generate the in-silico dataset, simulated biosignals are combined with a stochastic noise model, incorporating Gaussian noise, red noise at varying intensities, and random signal flipping, as further explained in \autoref{sec:simulations}.  We validate this measurement model through an ablation study, where we assess the robustness across different noise conditions of the NPE trained on: \textbf{1.} simulated APWs without added noise, \textbf{2.} simulated APWs with the proposed stochastic noise model, and \textbf{3.} simulated APWs with a fixed noise level equal to the highest one considered in the stochastic version. An example of the simulated biosignals is shown in \autoref{fig:APW_gen_at_several_noise}.
 The results are shown in \autoref{fig:synthetic_SNR_metrics} and indicate that the NPE model trained on the stochastic noise model achieves a good trade-off in terms of MAE and calibration across different SNR levels of the test measurements. Notably, the NPE trained on high noise levels under-performs on clean measurements and the one trained on clean data fails under noisy conditions. The NPE trained on stochastic noise, on the other hand, consistently demonstrates well-calibrated inverse solutions across nearly all noise levels, with only a slight degradation observed at negative SNR for CO regression. Thus, the rest of our discussion and experiments focuses on NPE trained on stochastic noise.

%These findings underscore that rigorous in-silico validation, as demonstrated here, is crucial for ensuring the robustness and reliability of the proposed approach before any application to real-world, in-vivo measurements. 

\paragraph{NPE enables comprehensive population-level uncertainty analyses.}

\begin{figure*}
    \centering
    \includegraphics[width=1.\textwidth]{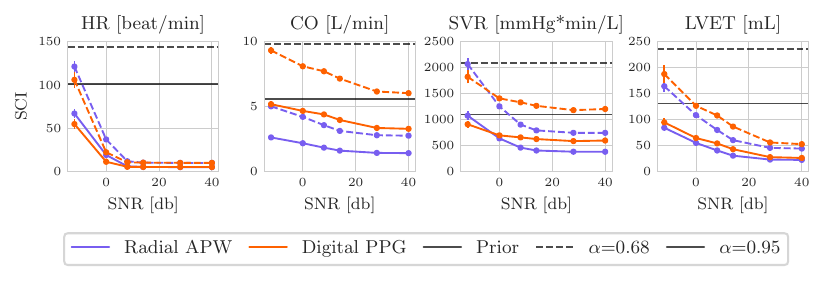} 
    \caption{Average SCI, for credibility levels $68\%$ and $95\%$. The x-axis denotes the SNR for both APWs and PPGs. %Results are averaged over five training instances, the vertical bars report one standard deviation. 
    % The units in square brackets are the physical units of the considered parameter. 
    }
    \label{fig:identifiability_analysis}
    \vspace{-1.5em}
\end{figure*}

With NPE, we leverage Bayesian inference to perform detailed uncertainty analysis on cardiac parameters of interest given each available biosignal.
In \autoref{fig:identifiability_analysis}, we show the average SCI derived from the biomarkers' posterior distribution, as learned by the NPE, to compare the informativeness of the radial APWs and of the digital PPGs for several noise levels. Given that the posterior distributions are well calibrated, as evidenced in \autoref{fig:synthetic_SNR_metrics}, the SCI directly relates to the precision of the predictive model and thus to the identifiability of the biomarker given the biosignal considered. %We can thus use the SCI to compare the informativeness of the radial APWs and of the digital PPGs for several noise levels.
% If the posterior estimators are calibrated, as evidenced in \autoref{fig:synthetic_SNR_metrics}, comparing their SCI against the prior distribution quantifies how much information a measurement carries about the biomarkers. 
Overall, the uncertainty in all parameters decreases significantly as the noise lowers, indicating that (a) the biosignals provide substantial information about these parameters, corroborating findings from other studies~\citep{melis2017bayesian, charlton2019modeling, charlton2022assessing}, and (b) the predicted uncertainty can serve as a reliable indicator of noisy samples, further validating the robustness of the NPE method. Moreover, the results underscore the unique information content of each measurement modality, with radial APWs offering more insights into CO and SVR compared to digital PPGs. The HR, on the other hand, is the most easily identified parameter across all measurements, except under very high noise conditions.

These findings illustrate that our framework facilitates an interpretable and comprehensive assessment of biomarker predictability from biosignals in-silico, which is crucial for guiding in-vivo experiments. 

%This in-silico study not only bridges the gap between theoretical and practical application but is an essential step toward translating these methods to clinical settings.

%% file: tex/results_invivo.tex
\subsection{In-vivo analysis}\label{sec:in-vivo}

In the previous section, we analyzed the performance of the proposed inference approach in-silico, focusing on the robustness of the method and on the identifiability of cardiac parameters of interest. While the NPE provides valuable insights and robust predictions, it is only through comparison with real-world data that we can validate them 
and ensure their applicability to clinical settings. For these reasons, we evaluate the proposed framework on the VitalDB dataset \cite{Lee2022VitalDBAH}, which contains in-vivo biosignals labeled with the measured biomarkers. In particular, the dataset contains recordings of APWs and PPGs, along with corresponding measurements of HR, CO, and SVR from non-cardiac surgery patients who underwent routine or emergency surgery. We perform pre-processing steps to ensure the quality and reliability of the data, and we divide the corresponding biosignals into $8$-second segments to match the time resolution used in the in-silico simulations. The processing steps and examples of such waveforms are showcased in \autoref{app:vitaldb}.
The patients and the corresponding biosignals are split into training, validation, and test set. The training and validation set consists of $25\text{ }086$ waveforms from $148$ subjects, and the test set includes $10\text{ }348$ waveforms from $49$ subjects. 
All results reported are on the test set, and error bars report the standard deviation over
five training instances. While the proposed pipeline trains the NPE solely on simulated data, the baselines and the hybrid learning strategy, which combine real and simulated data, use part of the training set.

\paragraph{Cardiac output monitoring of VitalDB patients.}

\begin{figure*}
    \centering
    \includegraphics[width=1.\textwidth]{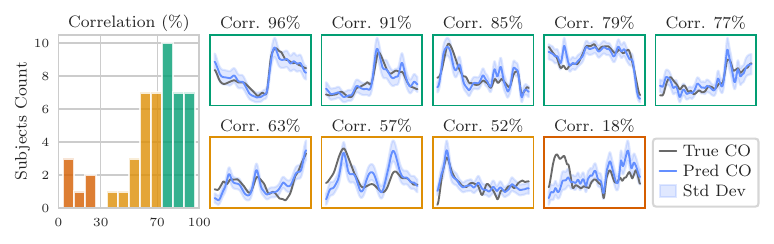}
    \caption{Left: histogram of the per-patient Spearman correlations between the measured CO in VitalDB and the predicted CO by NPE, which is trained solely on simulated APWs. Right: we plot the normalized CO tracking provided by VitalDB (True CO) together with the NPE prediction (Pred CO) for patients with different correlations.
    }
    \label{fig:tracking_vitaldb}
    \vspace{-1.5em}
\end{figure*}

Accurate monitoring of CO is of crucial importance in cardiovascular care. It is used, for example, to measure the effect of interventions \citep{Critchley2010ACR} and to assess an individual's cardiovascular efficiency, exercise capacity, and physical conditioning, making it also a valuable metric to track fitness~\citep{nystoriak2018cardiovascular, stickland2006does}. While intermittent pulmonary artery thermodilution and transpulmonary thermodilution are considered the gold standards for CO measurement, they are invasive and carry risks of significant complications~\citep{evans2009complications}. Consequently, numerous non-invasive CO monitoring devices have been developed; however, these devices are often less reliable, with error rates frequently exceeding the clinically acceptable threshold of $30\%$ \citep{Sanders2019AccuracyAP}. 

\begin{figure}[t!]
    \centering
    \includegraphics[width=1.\textwidth]{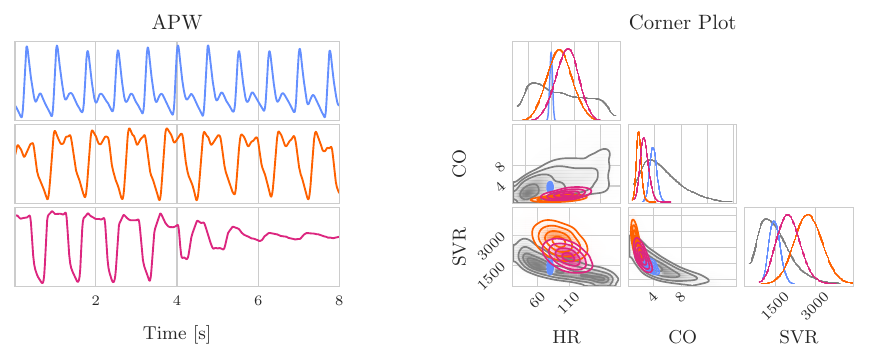}
    \caption{Uncertainty analysis of the NPE trained on synthetic APWs and tested on VitalDB APWs. Left: Three VitalDB APWs characterized by different cardiac functions and noise in the recording. Right: Corner plot showing the learned posterior distributions of the corresponding APWs and their prior distribution (grey).
    }
    \label{fig:uncertainty_vitaldb}
\end{figure}

Motivated by these limitations, we focus on evaluating the proposed framework’s performance in tracking the temporal trends of CO in VitalDB subjects using APWs as the primary measurement.
In \autoref{fig:tracking_vitaldb} (left), we compute the per-patient Spearman correlation between the CO recorded in VitalDB and the corresponding NPE prediction. As the posterior distributions are uni-modal (see \autoref{fig:uncertainty_vitaldb}), we use the expectation of the posterior distributions as point estimates.
Without access to any real-world measurements during training, the proposed framework demonstrates a strong ability in predicting accurate CO in the VitalDB test set. Remarkably, a majority of subjects exhibit a correlation greater than $70\%$ between the measured and predicted CO. These results highlight the potential of SBI to track CO trends even when trained exclusively on synthetic data.
Furthermore, \autoref{fig:tracking_vitaldb} (right) shows the normalized VitalDB CO monitoring through time for a random subset of patients, together with the NPE prediction.
The CO tracking predicted by our method captures the temporal trends of CO monitoring accurately given APWs as input. 

A similar analysis is conducted in \autoref{app:supp_results} for the tracking of SVR, which provides an additional tool for hypertension risk prediction and enables hemodynamic diagnosis in patients with acute congestive heart failure \citep{Khnen2020SystemicVR, cotter2003role}. The results demonstrate that the proposed method generalizes reasonably well for SVR tracking, although not as strongly as for CO. %, and therefore, it can accurately tracks different cardiac values for a significant number of subjects given APWs.

\paragraph{NPE enables per-individual uncertainty quantification.}

\begin{figure}[ht!]
    \centering
     % First subplot
    \begin{subfigure}[b]{\textwidth}
        \centering
        \includegraphics[width=\textwidth]{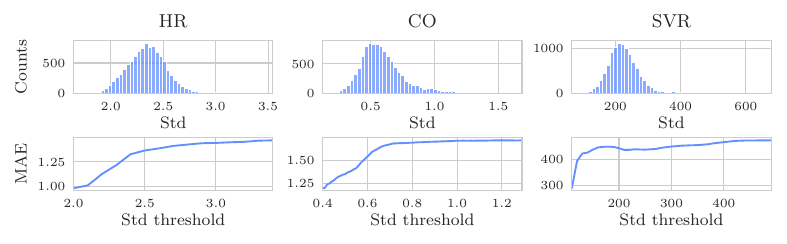} 
        \vspace{-2em}
        \caption{}\label{fig:std_mae}
    \end{subfigure}
    \begin{subfigure}[b]{\textwidth}
        \centering
        \includegraphics[width=1.\textwidth]{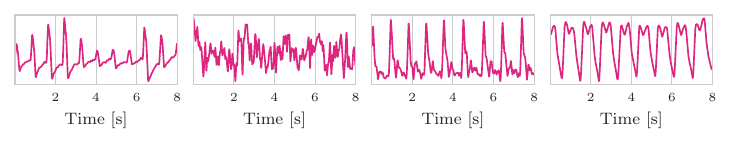}
        \vspace{-2em}
        \caption{}\label{fig:waveform_high_std}
    \end{subfigure}
    \caption{(a) Histograms of the standard deviations (Std) of the predicted posterior distributions of HR, CO, and SVR conditioned on VitalDB APWs and (b) random examples of APWs with corresponding high predicted HR uncertainty (i.e., high Std).}
\end{figure}

One of the key strengths of the proposed SBI-based framework is its ability to quantify uncertainty per individual measurement. This fine-grained analysis is crucial for understanding and addressing the variability inherent in physiological signals, particularly when dealing with noisy real-world data. The NPE method used in this study provides a consistent, multi-dimensional representation of uncertainty, allowing for detailed insights that extend beyond traditional uni-dimensional or population-aggregated sensitivity analyses. 

In this section, we extend our in-vivo evaluation of the NPE model trained on synthetic APWs. Unlike the previous section, our focus here is on analyzing the uncertainty of the proposed approach. Specifically, we investigate the learned posterior distributions of the HR, CO, and SVR when conditioned on real APWs from the VitalDB dataset.
In \autoref{fig:uncertainty_vitaldb}, we plot three APWs from VitalDB that have distinct morphologies and the corresponding corner plots, which depict the 1D and 2D projections of the posterior distributions, respectively as histograms and contour plots. The corner plots illustrate the prior distribution (in grey) and the three posterior distributions corresponding to the three VitalDB measurements. By investigating the learned posterior distributions, we have a clear visual representation of the uncertainty in the predictions, which highlights the dependencies and symmetries within the forward model. Specifically, the model predicts a high uncertainty for the orange and pink waveforms. We hypothesize that these waveforms correspond to pathological cases such as aortic regurgitation\cite{Chambers_Huang_Matthews_2019}, or to noisy measurements that have not been explored by the simulator.

In \autoref{fig:std_mae}, we depict the histograms showing the distribution of the standard deviations (Std) of the posterior distributions for the HR, CO, and SVR. We also plot the MAE between the measured and predicted biomarkers when restricting to samples with a standard deviation lower than a certain threshold. As the threshold for Std is lowered, the MAE decreases, confirming that filtering out uncertain predictions enhances predictive accuracy. 

Finally, in \autoref{fig:waveform_high_std}, we plot random examples of APWs from VitalDB that correspond to high predicted HR standard deviation. These examples show that high uncertainty in HR typically corresponds to very noisy measurements or measurements that the CV simulator fails to model, as hypothesized in the previous analysis.
These analyses underscore the value of providing a detailed, individualized representation of uncertainty, allowing for more personalized assessments of cardiovascular health and more robust predictions.

%TO PUT IN APPENDIX
% \figref{fig:npe_vs_laplace}  the estimation of uncertainty provided by NPE and Laplace's approximation~\citep{MacKay2003Information} around the expectation of the posterior distribution, which is representative of the underlying assumptions made in variance-based sensitivity analyses~(VBSAs). Similarly to VBSAs, Laplace's approximation models uncertainty through a second-order statistic over the population considered.
%Compared to Laplace's approximation, NPE yields tighter and better calibrated credibility intervals. Laplace's intervals tend to be overconfident for measurements that lead to multi-modal posterior distributions, and they are underconfident when the posterior is uni-modal. 
%Furthermore, a point estimator, even with Laplace's uncertainty estimation, will likely assign high density to low-density areas for certain observations, especially if the true posterior is multi-modal. Such inconsistent quantification of uncertainty may mislead downstream decisions. \appref{app:add_exp} showcases the $5$D posterior distributions corresponding to two test examples, which exhibit distinct uncertainty profiles and support further the necessity to use an expressive and observation-dependent quantification of uncertainty.

\subsection{Hybrid Learning mitigates model misspecification}\label{sec:hybrid_learning_results}

\begin{figure}[ht]
    \centering
    % First Row
    \begin{subfigure}[b]{0.95\textwidth}\includegraphics[width=1.\textwidth]{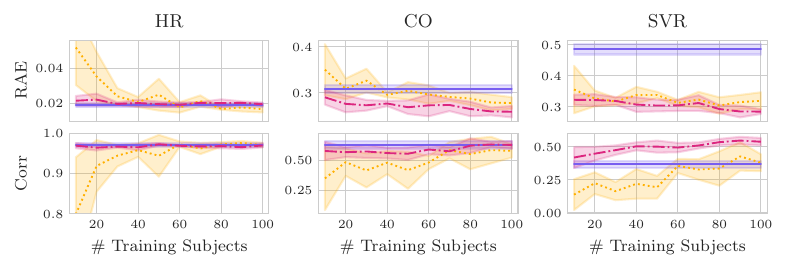}
    \caption{APW}\label{fig:APW_finetuning}
    % \vspace{-1.5em}
\end{subfigure}
    \hfill

\begin{subfigure}[b]{0.95\textwidth}\includegraphics[width=1.\textwidth]{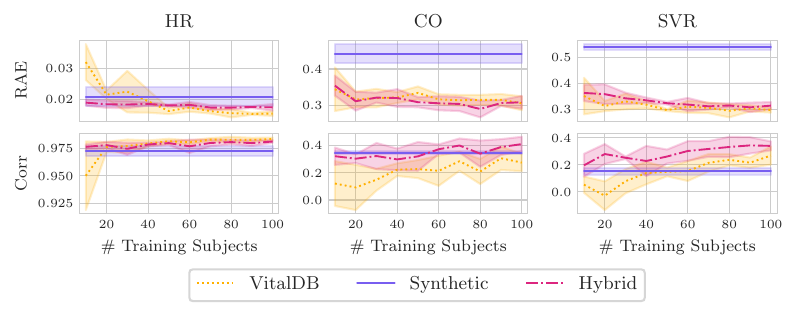}
    % \vspace{-1.5em}
    \caption{PPG}\label{fig:PPG_finetuning}
\end{subfigure}
\caption{RAE and per-patient correlation of various biomarkers tested on VitalDB using (a) APWs and (b) PPGs as biomarkers. We report the performance of NPE trained on synthetic data (Synthetic), NPE trained on synthetic data and finetuned on increasing number of VitalDB training subjects (x-axis) (Hybrid), and NPE only trained on increasing number of VitalDB training subjects (x-axis).
    }
\end{figure}

% \begin{figure*}
%     \centering
%     \includegraphics[width=1.\textwidth]{figures/PPG_finetuning.pdf}
%     \caption{Relative absolute error (RAE) and the average per patient correlation of various biomarkers tested on VitalDB using PPGs as measurements. We report the performance of NPE trained on synthetic PPGs (Synthetic), NPE trained on synthetic PPGs and finetuned on increasing number of VitalDB training subjects (x-axis) (Hybrid), and NPE only trained on increasing number of VitalDB training subjects (x-axis).
%     }
%     \label{fig:PPG_finetuning}
%     \vspace{-1.5em}
% \end{figure*}

In this section, we focus on the in-vivo  regression performance of the proposed approach in terms of the average Spearman correlation (Corr) per patient and the relative absolute error (RAE) between the predicted cardiac parameters and those reported in the VitalDB dataset. The latter can be directly compared to the $30\%$ relative accuracy of standard-of-care methods for estimating CO~\citep{Sanders2019AccuracyAP, bein2019best, park2016evaluation}.
The results are shown in \autoref{fig:APW_finetuning} for APWs and \autoref{fig:PPG_finetuning} for PPGs. The method, indicated as \emph{Synthetic} in the plots, shows low RAE and high correlation when predicting HR, indicating that SBI extends well to HR prediction in-vivo. In terms of CO regression, the RAE is also in line with the performance of current measurement devices \cite{Sanders2019AccuracyAP}, demonstrating that SBI can achieve reliable predictions for this critical biomarker. However, SVR shows higher bias and does not generalize as well. While this performance drop relative to CO is likely due to model misspecification, it is important to note that the "ground truth" SVR values in VitalDB are derived from the EV1000 system, whose accuracy in estimating SVR has been questioned~\citep{park2016evaluation}. This suggests that while SBI is effective for certain biomarkers, further refinement of the model may be necessary for others. Additionally, we observe a decrease in performance when using PPGs as biosignals compared to APWs, which we suspect stems from misspecification in the transfer function that converts pressure waveforms to PPGs. 

To mitigate the impact of model misspecification, we propose a hybrid learning strategy, detailed in \autoref{sec:hybrid_learning}, wherein we assume access to a small calibration set of \emph{labeled} in-vivo biosignals. The NPE trained on the synthetic dataset is subsequently fine-tuned using a weighted combination of real-world and synthetic biosignals. 
% During this process, only the encoder part of the network is finetuned, minimizing the risk of overfitting on the limited calibration set. 
In \autoref{fig:APW_finetuning} and \autoref{fig:PPG_finetuning}, the fully unsupervised NPE, trained solely on the synthetic dataset, is compared against (a) the hybrid learning strategy (\emph{Hybrid}), where finetuning is performed on a subset of VitalDB subjects, and (b) the fully supervised NPE trained entirely on VitalDB measurements (\emph{VitalDB}). The performance of the last two approaches is plotted against the number of VitalDB patients used for training/fine-tuning.

Interestingly, the synthetic approach often outperforms the fully supervised one in terms of correlation, suggesting that the synthetic data captures essential signal patterns. However, the method exhibits a higher bias, as indicated by increased RAE, especially in SVR regression and for PPG measurements. This underscores the limitations of relying solely on synthetic data when the absolute value of the predicted biomarker is of crucial importance. On the other hand, the hybrid learning strategy leverages the strengths of both in-silico and in-vivo data, effectively mitigating model misspecification. It achieves the best of both worlds as it outperforms both in-silico and in-vivo-only strategies in terms of correlation and MAE, with only a few in-vivo samples.

% In summary, by carefully examining the posterior distributions learned from simulations onto real-world data, we can identify aspects of the model that effectively transfer in-vivo and those that require further refinement. If necessary, in-silico analyses may guide the resolution of such misspecification by informing when and which parameters are hinted identifiable by the misspecified model. The next step is to gather the corresponding set of real-world labelled data and resolve the misspecification on this set, e.g. by modifying the noise model. A consecutive uncertainty analysis will tell if these parameters are still identifiable under the new and better-specified model.

% By stratifying the population based on expected levels of error using the posterior distribution, we can validate the transferability of the learned posterior distributions. Additionally, an important aspect left for future work is to propose new noise models that produces posterior distributions whose uncertainty is well-calibrated on real-world signals.

%% file: tex/discussions.tex
\section{Discussion}\label{sec:discussion}

\paragraph{Understanding Hemodynamics Simulators with SBI.}

% Studying the effect that cardiac parameters of interest have on the hemodynamics simulators is of crucial importance to better understand the complex cardiac models employed and to mitigate potential misspecification.
Building a new biomarker estimator requires a deep understanding of the causal mechanisms that underly measured signals and their sensitivity to both spurious and informative variation. To this end, variance-based sensitivity analysis~\citep[VBSA,][]{melis2017bayesian, piccioli2022effect,schafer2022uncertainty} has been often employed to explore the effects of various parameters of interest on the simulated measurements. 
We deviate from previous work and employ NPE, instead, an SBI approach that learns the posterior distributions of the parameters of interest given APWs or PPGs, and thus \emph{invert} the simulator. 
NPE-based uncertainty analysis supports the study of a richer set of uncertainty properties than VBSA. For instance, SBI quantifies uncertainty as a function of individual measurements, whereas VBSA provides a unique number for an entire population.
Furthermore, SBI enables the joint analyses of several biomarkers, while applying VBSA in this context is numerically challenging.
In the results section, we show that the representation of the solution's uncertainty of the proposed NPE approach successfully captures \textbf{(i)} the effect of nuisance parameters---parameters that ensure the model can describe diverse physiological conditions (e.g., arteries' elasticity, length and radii) and noise configurations, which are responsible for the forward model's stochasticity; 
% \textbf{(ii)} the symmetries of the forward model, leading to non-unique inverse solutions; 
\textbf{(ii)} the lack of sensitivity, magnifying small amount of measurement noise into high uncertainty on the estimated parameters; and \textbf{(iii)} the heterogeneity of the population considered, leading to uncertainty profiles that directly depend on the measured signal. We argue that SBI should become a standard tool to study complex cardiovascular simulators and better understand the properties of their inverse solutions, as we have demonstrated in \autoref{sec:in-silico}. Nevertheless, SBI's ability to inform about the real world is tailored to the simulator's faithfulness and can thus yield wrong conclusions in the presence of model misspecification. As an example, in \autoref{sec:in-vivo}, we observed that, while the NPE could track the temporal trend of CO in-vivo, it was unable to predict absolute values.

\paragraph{Misspecification of Cardiovascular Simulators.}

As highlighted by some of our empirical results, models are never a perfect representation of real-world data~\citep{box1976science}. Misspecification may hamper the practical relevance of insights extracted from a model~\citep {white1982maximum}. Thus, it is necessary to recognize and understand model misspecification to reason about the real world confidently~\citep{geweke2012prediction, box1976science}. Overcoming misspecification is most effectively achieved by identifying conclusions independent from the most critical sources of misspecification rather than illusively aiming for perfect models. For instance, in this work, the simulated waveforms strongly depend on the shape of the boundary inflow condition at the aorta, approximated with a simplistic and idealized five-parameter description (as detailed in \autoref{sec:methods}), clearly misspecified for some use cases. Nevertheless, this description accurately represents the relationship between a beat's length and the HR, and thus a predictive model of HR trained on in-silico data generalizes well to in-vivo measurements. Furthermore, our results validated less obvious qualities, such as the ability to describe the relationship between APWs and the relative values of CO and SVR. 

The foundational hypothesis of this work is that SBI can be used to capture relationships within the real data despite misspecifications inherently present in the simulator.  We then aim to demonstrate the potential of SBI to improve cardiovascular care by touching several steps of the scientific method process, which is typically composed of \textbf{1.} model analysis, \textbf{2.} real-world experimentation, \textbf{3.} comparison with observations, and \textbf{4.} model refinement. In \autoref{sec:in-silico}, we demonstrate its usefulness in extracting scientific hypotheses from the model (step \textbf{1.}), while evaluating the effect of different noise models on parameter identifiability. In this paper we do not tackle step \textbf{2.} directly, but instead rely on publicly available datasets suitable to prove the effectiveness of SBI on real-world data. In \autoref{sec:in-vivo}, we used SBI to discard noisy measurements automatically and to compare theoretical predictions implicitly made by the CV simulator and real-world data (step \textbf{3.}). Finally, in \autoref{sec:hybrid_learning_results}, we perform model refinement (step \textbf{4.}) and show that fine-tuning the SBI model on in-vivo data proves to be effective in improving the accuracy of biomarker prediction.

\paragraph{Next generation of CV simulators.}
Our empirical results demonstrate the value of full-body 1D hemodynamics models, e.g., for CO and SVR tracking from APW, but also highlight their current limitations for PPG modeling. In addition to addressing these issues, the next step for advancing these models is extending them beyond single waveform analysis to develop fully personalized subject profiles, incorporating temporal dynamics and individual physiological variations. Such personalized model would provide a foundation for accurately representing complex, multi-parameter interactions and enhancing our understanding of cardiovascular health.

In this context, future work should improve the simulation of PPGs, e.g., following a better understanding of the relationship between APWs and PPGs, as the non-invasive nature and broad availability of PPGs make them particularly relevant for long-term and continuous monitoring of cardiovascular health. Incorporating additional signals, such as electrocardiograms (ECGs), could further enhance this framework by providing insights into distinct yet interconnected physiological processes. For instance, ECG and PPG together shall enable the estimation of pulse transit time and indirectly provide information on the arteries stiffness properties and associated risks. 
In this context, traditional point estimators are limited in their capacity to combine information from multiple sources, whereas probabilistic frameworks allow for straightforward integration and do not necessarily need to build a joint forward model of the two modalities. Indeed, we can easily write the posterior distribution given the two measurements as the product of each posterior distribution: $p(\phi \mid x, y) \propto p(\phi \mid x) p(\phi \mid y)$ if we can assume conditional independence  $\mathcal{X} \perp \mathcal{Y} \mid \Phi$ between the two forward models $p(x\mid \phi)$ and $p(y\mid \phi)$, e.g., if $\phi$ is a common causal factor.

As cardiovascular models evolve and rely on additional parameters, they inevitably introduce new sources of uncertainty, further underlining the need for a probabilistic approach. SBI and machine learning offer powerful solutions for managing this complexity, allowing us to handle increased model dimensionality and stochasticity while providing interpretable posterior distributions. By using SBI methods, future research will be capable of effectively integrating multiple biosignal modalities and exploring the temporal dynamics of cardiovascular biomarkers—developments that would unlock better and more personalized cardiovascular care.

%% file: tex/method.tex
\section{Methods}\label{sec:methods}
% This section details the hemodynamics numerical simulation and formally the NPE algorithm and the metrics used to evaluate the inference results.

\subsection{Whole-body 1D Cardiovascular Simulations}\label{sec:simulations}

We define a simulator as a forward generative process $g: \Theta \rightarrow \mathcal{X}$ that inputs a vector of parameters $\mathbf{\theta} \in \Theta$ and returns a simulation $\mathbf{x} \in \mathcal{X}$.  
Our work utilizes the whole-body 1D cardiovascular simulator described in \cite{charlton2019modeling}, which models the hemodynamics in the $116$ largest human arteries, using the principle of mass and momentum conservation in blood circulation. The model's parameters describe the blood out-flowing the left ventricle, uni-dimensional physical properties of each artery, and a lumped-element model of the vascular beds.
%The full-body arterial model introduced in \cite{alastruey2012arterial}, on which \cite{charlton2019modeling} relies, describes the arterial pulse wave propagation into $116$ arterial segments, making up the largest arteries of the thorax, limbs, and head. 
This model is a good compromise between faithfulness to the real-world system and complexity~\citep{alastruey2023arterial}. It enables forward simulation of APWs and PPGs at multiple locations, given a set of physiological parameters describing the geometrical and physical properties of the cardiovascular system. %Running a simulation takes a few minutes on any standard CPU~\citep{melis2017bayesian}, allowing \cite{charlton2019modeling} to release a dataset of $4374$ simulated healthy individuals aged $25$ to $75$.
Compared to 3D and 0D models, 1D models offer a better balance between expressivity and efficiency. While 1D simulations may be less accurate than 3D models, they trade a modest and well-studied decrease in accuracy against much lighter simulation costs~\cite{xiao2014systematic, alastruey2023arterial}. Furthermore, the tractable parameterization and efficient simulation of 3D whole-body hemodynamics remain two open research questions~\citep{PEGOLOTTI2024107676, lasrado2022silico}. 
On the other side of the CV modeling spectrum, 0D simulations~\citep{john2004forward, shi2011review} rely on a lumped-element model to describe the relationship between blood flow at one location (e.g., left ventricle outflow) and blood pressure and flow at other locations. In addition to ignoring significant physical effects such as wave propagation and reflection, 0D models are partially parameterized by non-physiological quantities. Generating a representative population, such as the one considered in our study, can thus be challenging with these models.

\paragraph{Model description.}
Following \citep{alastruey2012arterial,charlton2019modeling}, we consider the compartmentalized arterial model made of the following sub-models: \textbf{1.} the heart function; \textbf{2.} the arterial system; \textbf{3.} the geometry of arterial segments; \textbf{4.} the blood flow; and \textbf{5.} the vascular beds. \textit{The heart function} describes the blood volume along time at the aorta as a five-parameter function. \textit{The arterial system} is described as a graph, the heart is the parent root, and then arteries branch out into the body. Every branch of the network represents an arterial segment. Segments are coupled so that the conservation of mass and momentum hold in the complete system. Additionally, the heart function defines the boundary condition on the parent root of the arterial network. The vascular bed describes the boundary condition on the leaf nodes. \textit{The geometry of arterial segments} assumes the segments are axial-symmetric and tapered tubes. Hence, the geometry of each arterial segment can be described using 1D parameters such as radius and thickness of the arterial wall. \textit{The blood flow} in the 1D segments follows fluid dynamics, which depends on the geometry and visco-elastic properties of the arterial wall. \textit{The vascular beds} are modeled using 0D approximations, i.e., the geometrical description is being lumped into a space-independent parametric transfer function. 

% Following this high-level overview of the 1D hemodynamics models, we now provide some mathematical details on the core components. 
The main state parameters of whole-body 1D hemodynamics models are the volumetric flow rate $Q(z,t)$, the blood pressure $P(z, t)$, and the vessel cross-sectional area $A(z,t)$ at axial position $z$ and time $t$, in each artery considered. Based on the conservation of mass and momentum, one can derive the partial differential equations (PDEs)
\begin{align}
    \frac{\partial A}{\partial t} + \frac{\partial Q}{\partial z} &= 0 \\
    \frac{\partial Q}{\partial t} + \frac{\partial }{\partial z} \left(  \alpha \frac{Q^2}{A}\right) + \frac{A}{\rho} \frac{\partial P}{\partial z} &= -2 \frac{\mu}{\rho} (\gamma_\nu +2)\frac{Q}{A},
\end{align}
where $\alpha$ is the Coriolis' coefficient, $\mu$ is the blood dynamic viscosity, and $\gamma_\nu$ is a parameter defining the shape of the radial velocity profile. A third relationship of the arterial wall mechanics relates pressure and cross-section area as
\begin{align}
    P(A) = P_{ext} + \beta \left( \sqrt{A} - \sqrt{A_0}\right)
    + \frac{\Gamma}{\sqrt{A}} \frac{\partial A}{\partial t}, \\
     \text{where } \beta = \frac{4}{3} \frac{\sqrt{\pi} Eh_0}{A_0} \text{ and } \Gamma = \frac{2}{3} \frac{\sqrt{\pi} \varphi h_0}{A_0}
\end{align}
respectively denote the elastic and viscous components of the Voigt-type visco-elastic tube law,
$P_{ext}$ is the reference pressure at which the geometry is described by the cross-sectional area $A_0$ and thickness of the arterial wall $h_0$. The elastic modulus $E$ and wall viscosity $\varphi$ characterize the mechanical properties of the wall. In addition to these PDEs, boundary conditions are formulated by coupling each artery segment with the parents and children in the arterial network. For further details, see \cite{Melis2017, charlton2019modeling, alastruey2012arterial}.

\paragraph{In-silico Dataset Generation}
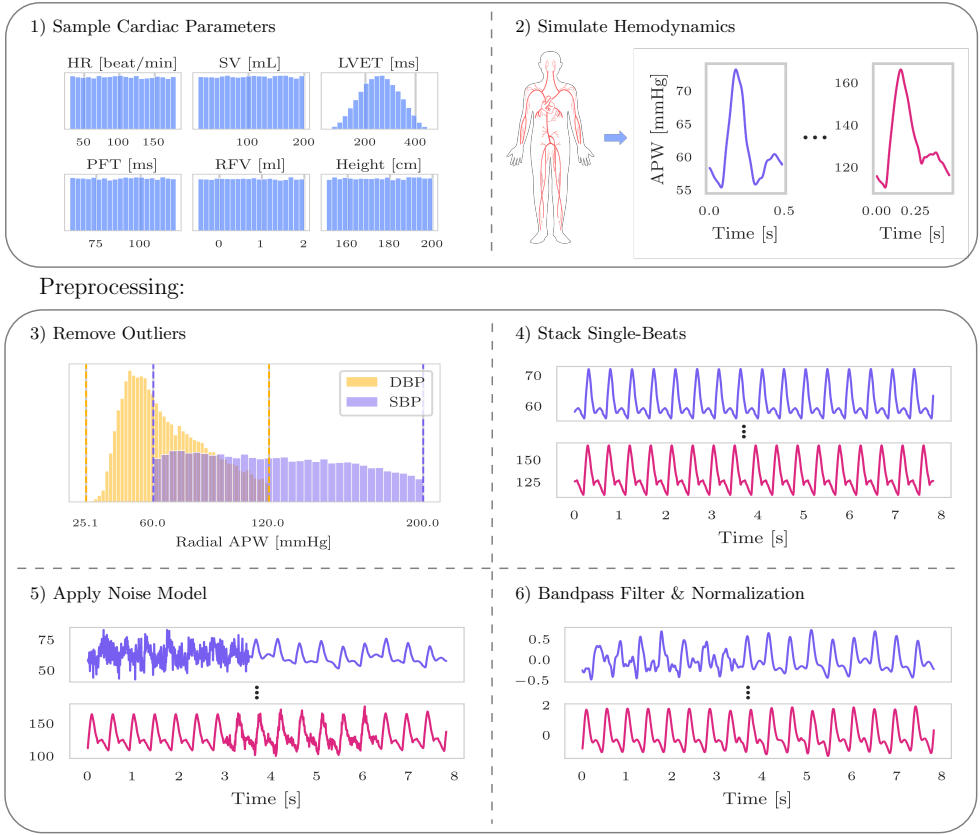
\begin{figure}[ht]
    \centering
\input{figures/tikz/generation_figure}
  \caption{Generation of radial APWs and digital PPGs in-silico. 1) Cardiac parameters are sampled randomly; 2) single-beat APWs at random radial artery locations and PPGs at random digital arteries are simulated using whole-body 1D cardiovascular simulation and 1D blood flow solver based on MUSCL finite-volume numerical scheme; 3) waveforms with out-of-range DPB and SBP are filtered out; 4) the same beats are concatenated multiple times and the resulting time-series are randomly cropped into 8-second segments; 5) an amortized stochastic noise model is applied to the resulting APWs and PPGs; 6) the waveforms are filtered using the Butterworth second-order bandpass filter.
  }\label{fig:pre-processing}
    \vspace{-1.5em}
\end{figure}

Given the forward model described above, we construct a comprehensive dataset\footnote{The dataset will be released upon acceptance.} of hemodynamic simulations that provides uniform coverage of cardiovascular parameters. %This dataset is designed to support both in-silico and in-vivo analyses for accurate biomarker estimation from biosignals.
We consider a population aged $25$ to $75$ with several free parameters $\theta$ that model heterogeneous cardiac, arterial, and vascular bed properties (see Appendix \ref{app:hemodynamics_model} for the complete list). We use the arterial and vascular beds properties defined in \cite{charlton2019modeling},
where we modify the length of the arterial segments ($l_{a}$) by assuming a linear relationship with the height (h) of the subject: $l_{a^{*}}= l_{a} \times\frac{h + \epsilon}{170}$, where $\epsilon \sim \mathcal{N}(0, 5)$.
We then sample the heart function properties—Heart Rate (HR), Stroke Volume (SV), Peak Flow Time (PFT), and Reverse Flow Volume (RFV)— as well as the height (h) from uniform distributions, creating $80'000$ virtual subjects (Step $1$ in Fig. \ref{fig:pre-processing}). The Left Ventricular Ejection Time, LVET, is derived from the empirical relationship defined in \cite{charlton2019modeling}, where we introduce stochasticity and noise to address and mitigate potential model misspecification: 
% $$LVET = (244 \pm 40) - (0.926 \pm 0.05) HR + (1.08\pm0.05)SV.$$
$$LVET = (244 + \epsilon_1) - (0.926 + \epsilon_2) HR + (1.08+ \epsilon_3)SV,$$
where $\epsilon_i$ are sampled independently for each subject from uniform distributions, specifically $\epsilon_1 \sim \mathcal{U}_{[-40, 40]}$ and $\epsilon_2, \epsilon_3 \sim \mathcal{U}_{[-0.05, 0.05]}$.

% The measurements considered in this study are signals commonly collected in intensive care units (ICUs): the arterial pressure waveform (APW) at the radial arteries and the photoplethysmogram (PPG) at the digital arteries. We generate a database\footnote{The dataset is publicly available at ...} of hemodynamic simulations following the steps showcase in Fig. \ref{fig:pre-processing}.
For each virtual subject, we solve the corresponding partial differential equations, using the open-source 1D blood flow solver based on MUSCL finite-volume numerical scheme from \citep{openBF.jl-2018}. We generate single heartbeat APWs from the radial artery, where we randomly sample both the arm (either left or right) and the exact location in the artery to increase variability in the simulations (Step $2$ in Fig. \ref{fig:pre-processing}). PPGs at the digital arteries are instead simulated by normalizing the pulsatile variation in the volume of blood stored in the terminal Windkessel model \cite{charlton2019modeling}.
In our experiments, we filter out APWs and corresponding PPGs with anomalous amplitude ranges to ensure that the distribution of systolic and diastolic blood pressures (SBP and DBP) matches real-world observations (Step $3$ in Fig. \ref{fig:pre-processing}). Namely, we reject APWs if $\text{DBP} >  120 \text{mmHg}$ or $\text{SBP} \notin [60 \text{mmHg}, 200 \text{mmHg}]$. The number of virtual patients after filtering is $32'246$.
The resulting APWs and PPGs are further pre-processed by concatenating the same beat multiple times and randomly cropping the time-series into 8-second segments (Step $4$ in Fig. \ref{fig:pre-processing}). This step ensures that all potential starting positions within a heartbeat are accounted for.

To bridge the gap between simulated and real-world data, we apply a stochastic measurement model  (Step $5$ in Fig. \ref{fig:pre-processing}), incorporating three noise models: Gaussian noise, Red noise, and flipping. The Gaussian and Red noise are amortized by sampling random intensity from an Exponential distribution, which has decreasing probabilities for higher noise intensity.  These two random noises are applied to random segments of $80\%$ of the waveforms, while flipping is applied to $30\%$ of the waveforms. 
The selected amortized noise model is designed to make the analysis less sensitive to model misspecification and noise in the measurements.
Finally, the waveforms are filtered using a Butterworth second-order bandpass filter with a filtering range of $0.5$–$10$ Hz \citep{zhangfiltering2021} (Step $6$ in Fig. \ref{fig:pre-processing}).

The resulting simulations, illustrated in the Appendix, \autoref{app:fig:synthetic_apw} and \autoref{app:fig:synthetic_ppg},
effectively capture the heterogeneity present in real-world data across different individuals, biosignals, and measurement conditions. %In comparison, current synthetic public datasets \cite{charlton2019modeling} only cover healthy distribution of cardiac properties and are based on coarse grids of cardiac parameters values, resulting in unrealistic multimodal-modal prior distributions that do not generalize well to real-world observations, as demonstrated in the Appendix (TODO).

% TO ADD: Inital number of apws: 81660, after filtering: 32246

\subsection{Simulation-based inference} \label{sec:sbi}
\label{sec:SBI}
%Faithful simulators are usually stochastic, for instance when they combine together a noise model that represents measurement errors with a deterministic mechanistic model that relates parameters to measurable quantities, such as the PDE model discussed before. Another common source of randomness are nuisance parameters. Indeed, we often distinguish between the parameters of interest $\mathbf{\phi} \in \Phi \subset \Theta $ and the nuisance parameters $\mathbf{\psi} \in \Psi \triangleq \Theta \setminus \Phi$ that are necessary to run the simulations but not of direct interest for downstream tasks. We usually aim to marginalise out the effect of those nuisance parameters by randomising their value following a marginal distribution $p(\mathbf{\psi})$, which transmits uncertainty to the simulation output $\mathbf{x}$ given a parameter of interest $\mathbf{\phi}$. These multiple sources of randomness call for a statistical treatment of inference over simulations~\citep{cranmer2020frontier, brehmer2020mining, cranmer2015approximating}. 
Complex simulators, such as the one described in the previous section, are poorly suited for inference and lead to intractable inverse problems. Let us divide the input parameters $\mathbf{\theta} \in \Theta$ into parameters of interest $\mathbf{\phi} \in \Phi \subset \Theta $ and nuisance parameters $\mathbf{\psi} \in \Psi \triangleq \Theta \setminus \Phi$ that are necessary to run the simulations but not of direct interest for downstream tasks (e.g., the lengths and radii of arteries). The inverse problem aims at estimating the parameters $\mathbf{\phi}$ given the simulations $\mathbf{x}$.
%We can abstract all sources of randomness within the nuisance parameters $\psi$ (e.g., even including the stochasticity of the noise model) such that the function $g: \Psi \times \Phi \rightarrow \mathcal{X}$ becomes deterministic. The deterministic simulator together with the sampling of nuisance parameters implicitly defines a likelihood function over the parameters of interest as 
%$p(\mathbf{x} \mid \phi) = \int \delta_{\mathbf{x}}\bigl( g\left(\psi, \phi \right) \bigr) p(\psi) d \psi, $ where $\delta_{\mathbf{x}}$ is the Dirac delta distribution centred at $\mathbf{x}$. 
In most cases, including ours, one can sample from the corresponding distribution $p(\mathbf{x} \mid \phi)$, but not evaluate it. Indeed, the latter operation requires nuisance parameters marginalization, which is intractable in most cases. The intractability of the likelihood function makes statistical inference over such models challenging.
In this context, SBI provides efficient tools to perform likelihood-free statistical inference.

% P1 defines what is SBI
% P2 defines normalizing flows to model distributions
\paragraph{Neural Posterior Estimation (NPE).}
% We take a Bayesian approach and assume that a well-motivated prior distribution $p(\phi)$ over the parameters is given. We rely on NPE~\citep{papamakarios2016fast, lueckmann2017flexible}, a popular SBI algorithm. 
SBI algorithms are broadly categorized as \textit{Bayesian} vs \textit{frequentist} and \textit{amortized} vs \textit{online} methods. In contrast to frequentist methods that only assume a domain for the parameter values, Bayesian methods rely on a prior distribution that encodes a-priori belief and target the posterior distribution, modeling the updated belief on the true parameter value after making an observation. As the prior distribution gets uninformative or the number of observations grows, both methods eventually yield the same inference results, the remaining difference being the interpretation of probabilities as a representation of belief or of intrinsic stochasticity. While online methods focus on a particular instance of the inference problem, i.e., with a fixed measurement value $\mathbf{x}$, amortized methods directly target a surrogate of the likelihood~\citep{vandegar2021neural}, the likelihood ratio~\citep{cranmer2015approximating, hermans2020likelihood}, or the posterior~\citep{lueckmann2017flexible,papamakarios2016fast} that is valid for all possible observations. 

Amortized methods are particularly appealing when repeated inference is required, as new observations can be processed efficiently after an offline simulation and training phase. In this work, we also perform repeated inference over the population of interest and have access to a properly defined prior distribution. Thus, we rely on neural posterior estimation~\citep[NPE,][]{papamakarios2016fast, lueckmann2017flexible}, a \textit{Bayesian} and \textit{amortized} method, which learns a surrogate of the posterior distribution $p(\phi \mid \mathbf{x})$ with conditional density estimation.
Within this framework, we train a parametric conditional density estimator for the parameters of interest $p_{\omega}(\phi \mid \mathbf{x})$ on the in-silico dataset $\mathcal{D} := \{(\phi_i, \mathbf{x}_i)\}_{i=1}^N$, composed of samples from the joint distribution $p(\phi, \mathbf{x}) =  \int p(\phi, \psi) p(\mathbf{x}\mid \phi, \psi) d\psi$. In this work, we rely on a rich class of neural density estimators called normalizing flows~\citep[NF, ][]{tabak2010density, tabak2013family, rezende2015variational, kobyzev2020normalizing, papamakarios2021normalizing}, from which both density evaluation and sampling is possible. 

% The NPE algorithm optimises the parameters of the neural density estimator with stochastic gradient descent on the Kullback-Leibler divergence between the conditional density estimator and the posterior distribution. 
Given an expressive class of neural density estimators $\{p_\omega(\phi \mid \mathbf{x}): \omega \in \Omega\}$, NPE aims to learn an amortized posterior distribution $p_{\omega^\star}(\phi \mid \mathbf{x})$ that works well for all possible observations $\mathbf{x} \in \mathcal{X}$, by solving
\begin{align}
    % & \omega^\star \in \arg\min_{\omega \in \Omega} \mathbb{KL}\left[ p(\phi \mid \mathbf{x}) \parallel p_\omega(\phi \mid \mathbf{x}) \right] \quad \forall \mathbf{x} \in \mathcal{X}  \label{eq:goal_NPE}\\
     & \omega^\star \in \arg\min_{\omega \in \Omega} \mathbb{E}_{\mathbf{x}} \left[ \mathbb{KL}\left[ p(\phi \mid \mathbf{x}) \parallel p_\omega(\phi \mid \mathbf{x}) \right] \right] \label{eq:expectation}\\ 
     \iff & \omega^\star \in \arg\min_{\omega \in \Omega} \int p(\mathbf{x}) p(\phi \mid \mathbf{x}) \left[ \log \frac{p(\phi \mid \mathbf{x})}{ p_\omega(\phi \mid \mathbf{x})} \right] d\mathbf{x} d\phi\\
    \iff  & \omega^\star \in \arg\max_{\omega \in \Omega} \int p(\mathbf{x}) p(\phi \mid \mathbf{x}) \log p_\omega(\phi \mid \mathbf{x}) d\mathbf{x} d\phi\\
    \iff  & \omega^\star \in \arg\max_{\omega \in \Omega} \mathbb{E}_{(\phi, \mathbf{x})} \left[ \log p_\omega(\phi \mid \mathbf{x}) \right]. \label{eq:objective_NPE}
    \end{align}
% \eqref{eq:goal_NPE} indicates the goal of NPE: learning a surrogate $p_{\omega^\star}$ that equals the posterior distribution for all values $\mathbf{x} \in \mathcal{X}$. Under the assumption that the class of functions considered contains the true posterior $p(\phi\mid \mathbf{x})$, the minimisers $\omega^\star$ of \eqref{eq:goal_NPE} and \eqref{eq:expectation} are the same. 

In practice, NPE approximates the expectation in \eqref{eq:objective_NPE} with an empirical average over the training set $\mathcal{D}$ and relies on stochastic gradient descent to solve the corresponding optimization problem. Assuming $\phi \in \mathbb{R}^k$ and unpacking the evaluation of the NF-based conditional density estimator, the training loss is
\begin{align}
    \ell(\mathcal{D}, \omega) 
    % &= \frac{1}{N}\sum_{i=1}^N \log p_{\omega}(\phi_i \mid \mathbf{x}_i)\\
    &= \frac{1}{N}\sum_{i=1}^N \log p_z\biggl(f_{\omega}\bigl(\phi_i; \mathbf{x}_i\bigr)\biggr) + \log \lvert J_{f_\omega}(\phi_i; \mathbf{x}_i) \rvert, \label{eq:loss}
\end{align}
following from the change-of-variables theorem~\citep{tabak2013family}.
The symbol $p_z$ denotes the density function of an arbitrary $k$-dimensional distribution (e.g., an isotropic Gaussian), $f_\omega:\mathbb{R}^k \times  \mathbb{R}^D\rightarrow \mathbb{R}^k$ denotes a continuous function invertible for its first argument $\phi$, parameterized by a neural network, and $\lvert J_{f_\omega} \rvert$ denotes the absolute value of the Jacobian's determinant of $f_\omega$ with respect to its first argument. In addition to density evaluation, as in \eqref{eq:loss}, the NF enables sampling from the modeled distribution by inverting the function $f_\omega$. 

In this work, we decompose $f_{\omega}\bigl(\phi_i; \mathbf{x}_i\bigr)$ into $ f_\gamma\bigl(\phi_i; f_\nu(\mathbf{x}_i)\bigr)$, where $f_\nu: \mathbb{R}^D\rightarrow \mathbb{R}^M$ is a convolutional neural network that encodes the observations into a lower dimensional representation, while $f_\gamma:\mathbb{R}^k \times \mathbb{R}^M\rightarrow \mathbb{R}^k$ is the invertible function with respect to its first argument, parameterized as a three-step autoregressive affine NF~\citep{papamakarios2017masked}, which offers a good balance between expressivity and sampling efficiency as demonstrated in \cite{wehenkel2020you}. 
% These models have an inductive bias towards simple density functions~\citep{verine2023expressivity}, which support that the multi-modality and diversity of posterior distributions observed in the population is not an artifact of our analysis but follows from the 1D cardiovascular model and prior considered. 
We provide additional details on the parameterization of $f$ and the sampling algorithm in \appref{app:NF}.

\paragraph{Hybrid Learning.} 
\label{sec:hybrid_learning}
The simulator described in \autoref{sec:simulations} offers a simplified representation of the complex interactions between cardiovascular parameters and noise models found in real-world data. Consequently, it is preferable to align the model’s predictions with real-world distributions. In this study, we refine the model by fine-tuning the NPE method using a small calibration set, $\mathcal{D}_C$, which is composed of $N_C$ real-world \emph{labeled} observations.
Specifically, after training the model by minimizing \autoref{eq:loss} on the in-silico dataset $\mathcal{D}$, the model is further fine-tuned with an objective function that weights equally the original loss function from \eqref{eq:loss} on synthetic data and on labeled observations from the calibration set, i.e., 
\begin{align}
    \tilde{\ell}(\mathcal{D}_C, \mathcal{D}, \omega) 
    &= \frac{1}{N_C}\sum_{i=1}^{N_C} \log p_z\biggl(f_\gamma\bigl(\phi_i; f_\nu(\mathbf{x}_i)\bigr)\biggr) + \log \lvert J_{f_\gamma}\bigl(\phi_i; f_\nu(\mathbf{x}_i)\bigr) \rvert + \ell(\mathcal{D}, \omega).\label{eq:loss_finetune}
\end{align}
% where $\mathcal{F} = \mathcal{D}_C + \mathcal{D}_S$. Our fine-tuning strategy involves a re-weighting scheme that ensures both in-silico and in-vivo measurements are equally represented to avoid overfitting on the few real-world examples, resulting in $\mathcal{D}_S \subset \mathcal{D}$ with $\lvert \mathcal{D}_S \lvert = N_C$ chosen at random at each training epoch.
During optimization, the parameters $\gamma$ of the three-step autoregressive affine NF are kept frozen to retain the statistical relationships established during in-silico training, while $f_\nu$, parametrized by the CNN, is refined. This approach aims to mitigate the distributional shift between in-vivo and in-silico measurements.
Experimental results demonstrate that this fine-tuning approach preserves the generalization capabilities of the model trained in-silico while significantly enhancing its accuracy on real-world observations.

%% file: figures/tikz/generation_figure.tex
\resizebox{\linewidth}{!}{

\tikzset{every picture/.style={line width=0.75pt}} %set default line width to 0.75pt        

\begin{tikzpicture}[x=0.75pt,y=0.75pt,yscale=-1,xscale=1]
%uncomment if require: \path (0,635); %set diagram left start at 0, and has height of 635

%Image [id:dp9142060785329923] 
\draw (494.5,123.56) node  {\includegraphics[width=233.62pt,height=107.17pt]{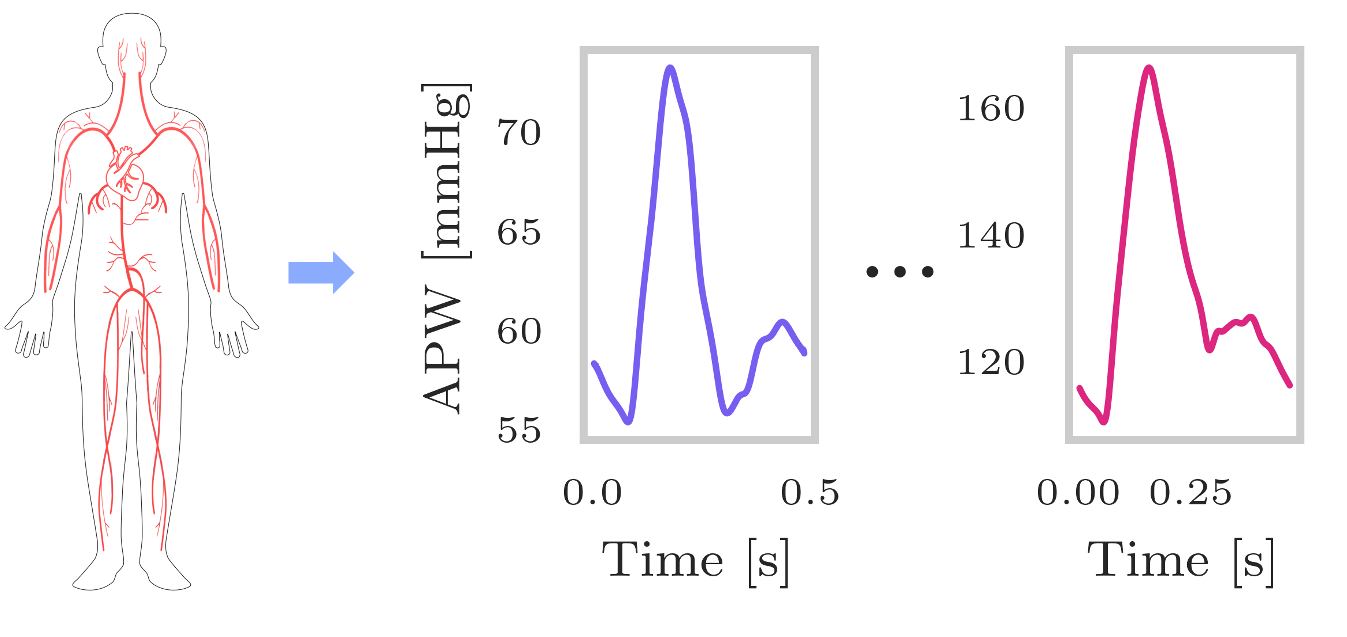}};
%Image [id:dp08861315558372651] 
\draw (165.13,506.56) node  {\includegraphics[width=235.69pt,height=107.17pt]{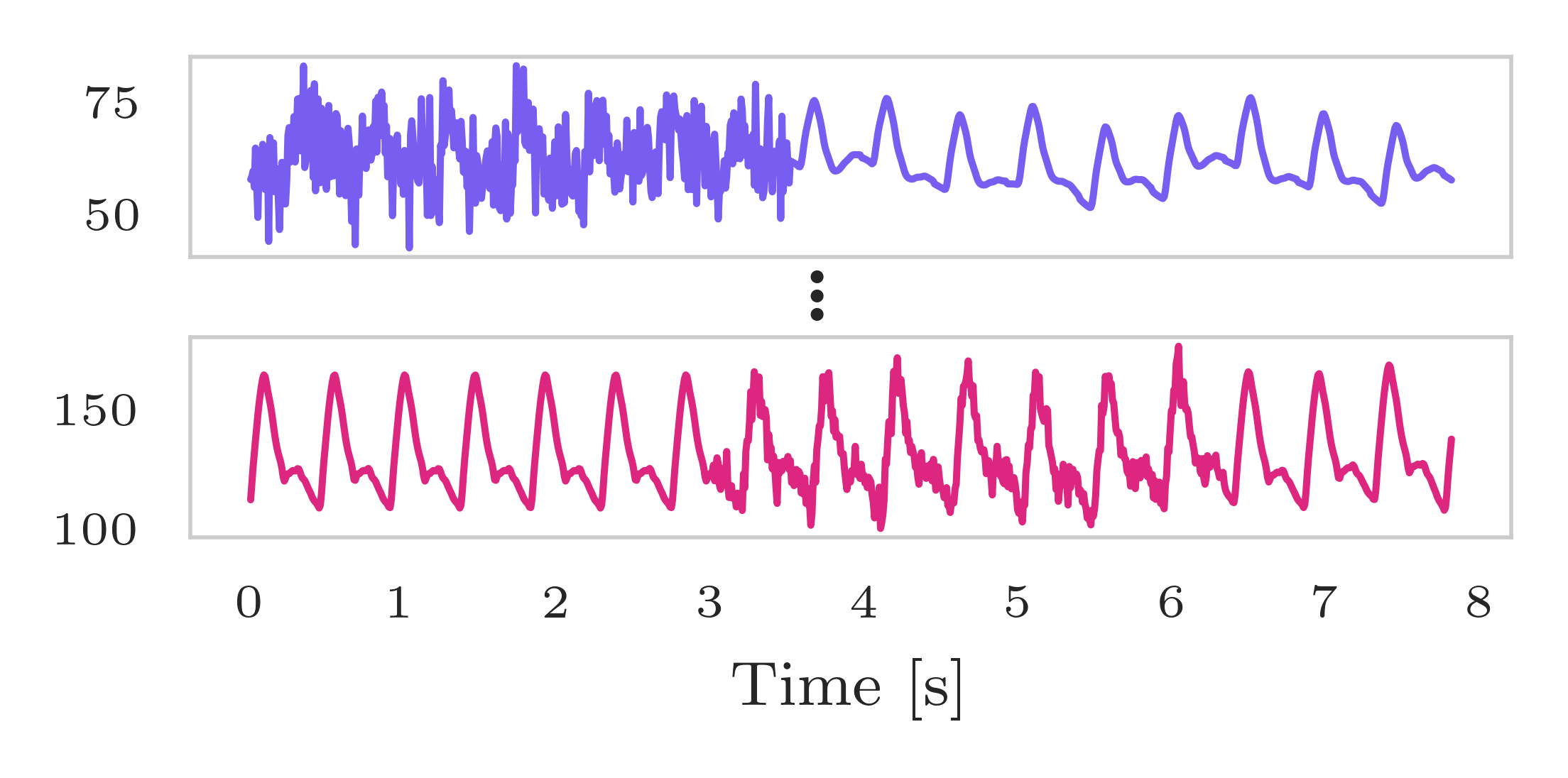}};
%Image [id:dp6447122747739202] 
\draw (492.5,329.5) node  {\includegraphics[width=235.12pt,height=107.25pt]{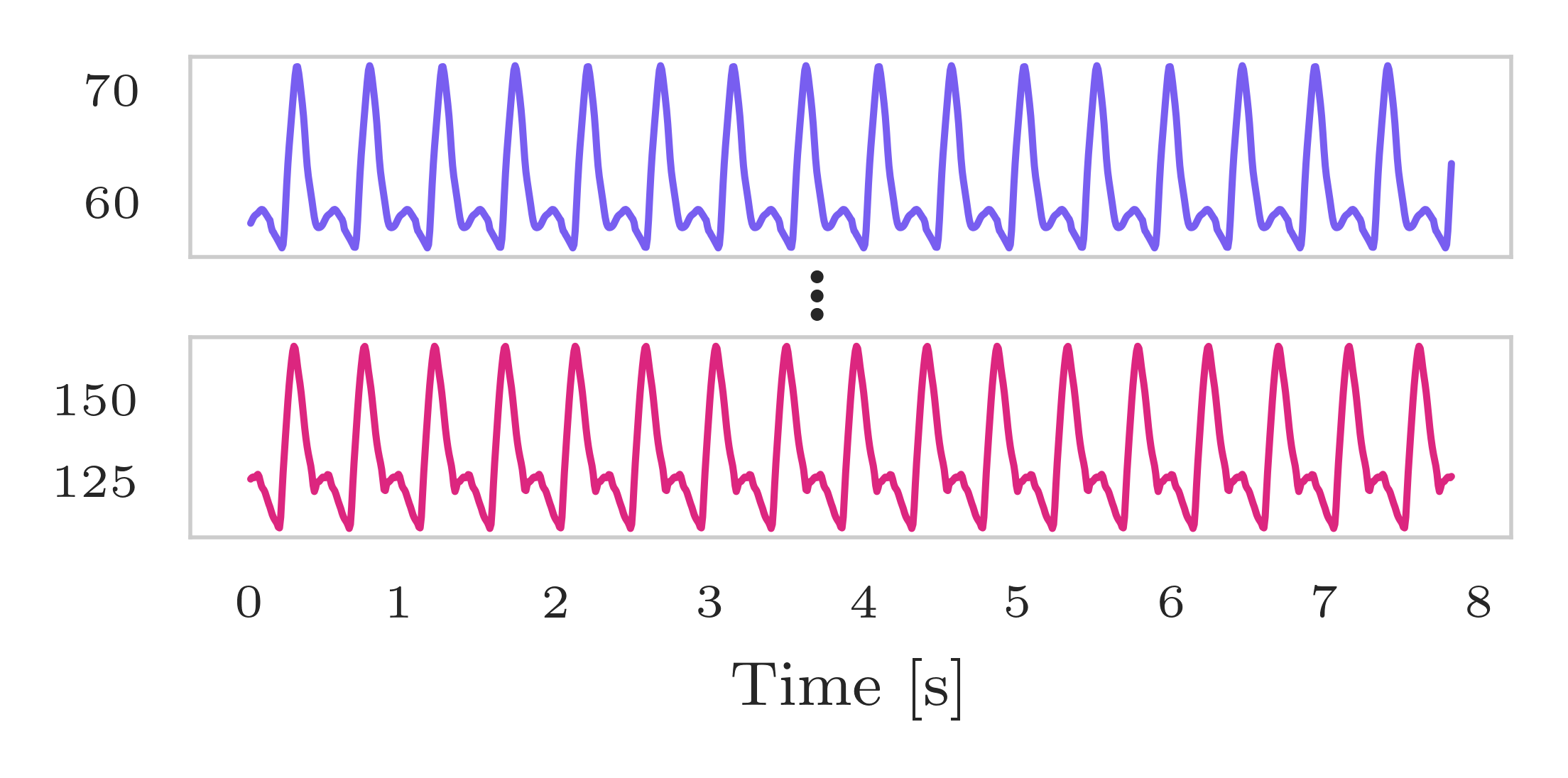}};
%Image [id:dp5285710560457696] 
\draw (492.13,506.56) node  {\includegraphics[width=235.69pt,height=107.17pt]{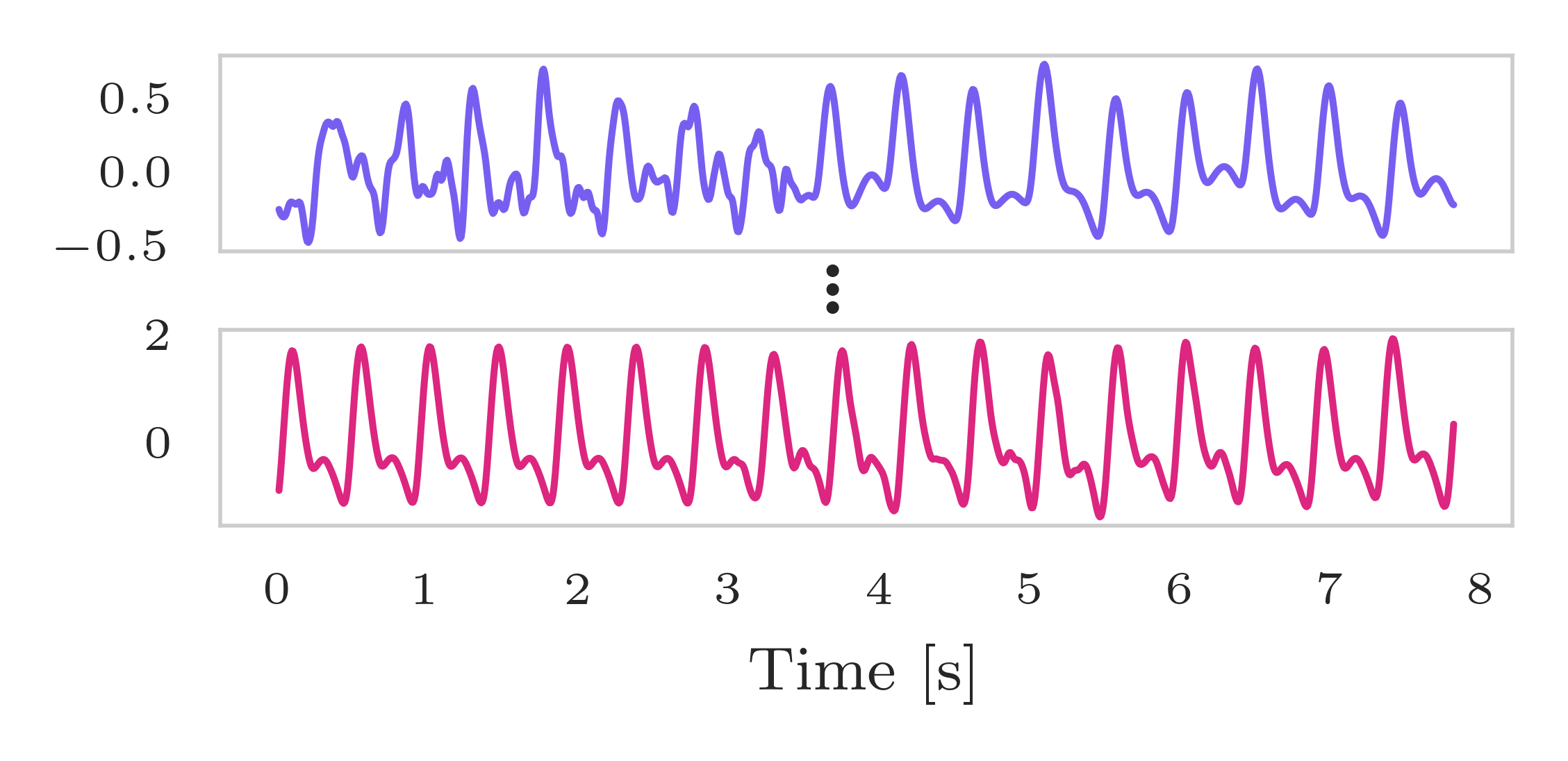}};
%Image [id:dp8024734296500414] 
\draw (170,123.56) node  {\includegraphics[width=204.94pt,height=110.17pt]{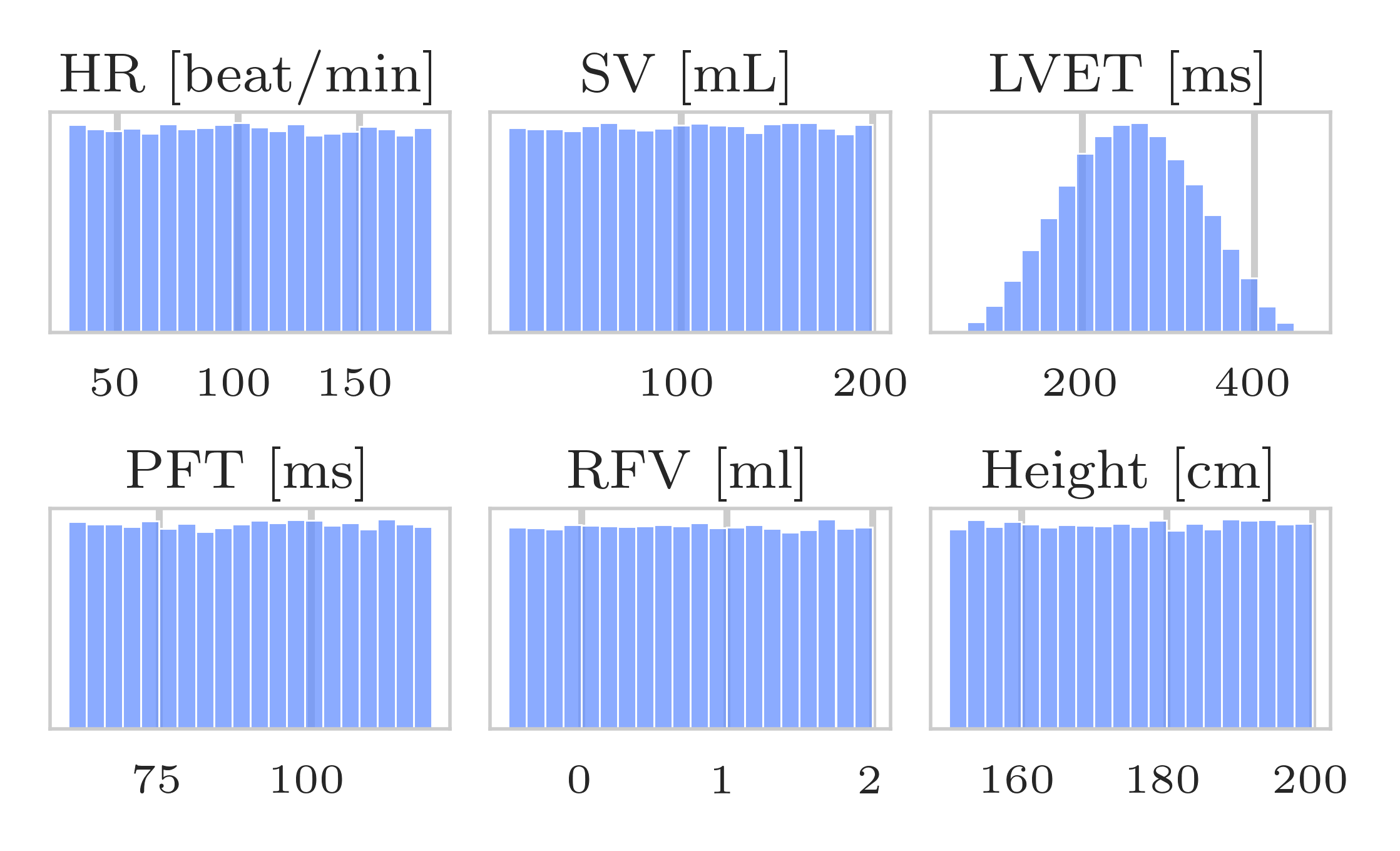}};
%Image [id:dp9382032254192376] 
\draw (170,329.5) node  {\includegraphics[width=200.56pt,height=107.25pt]{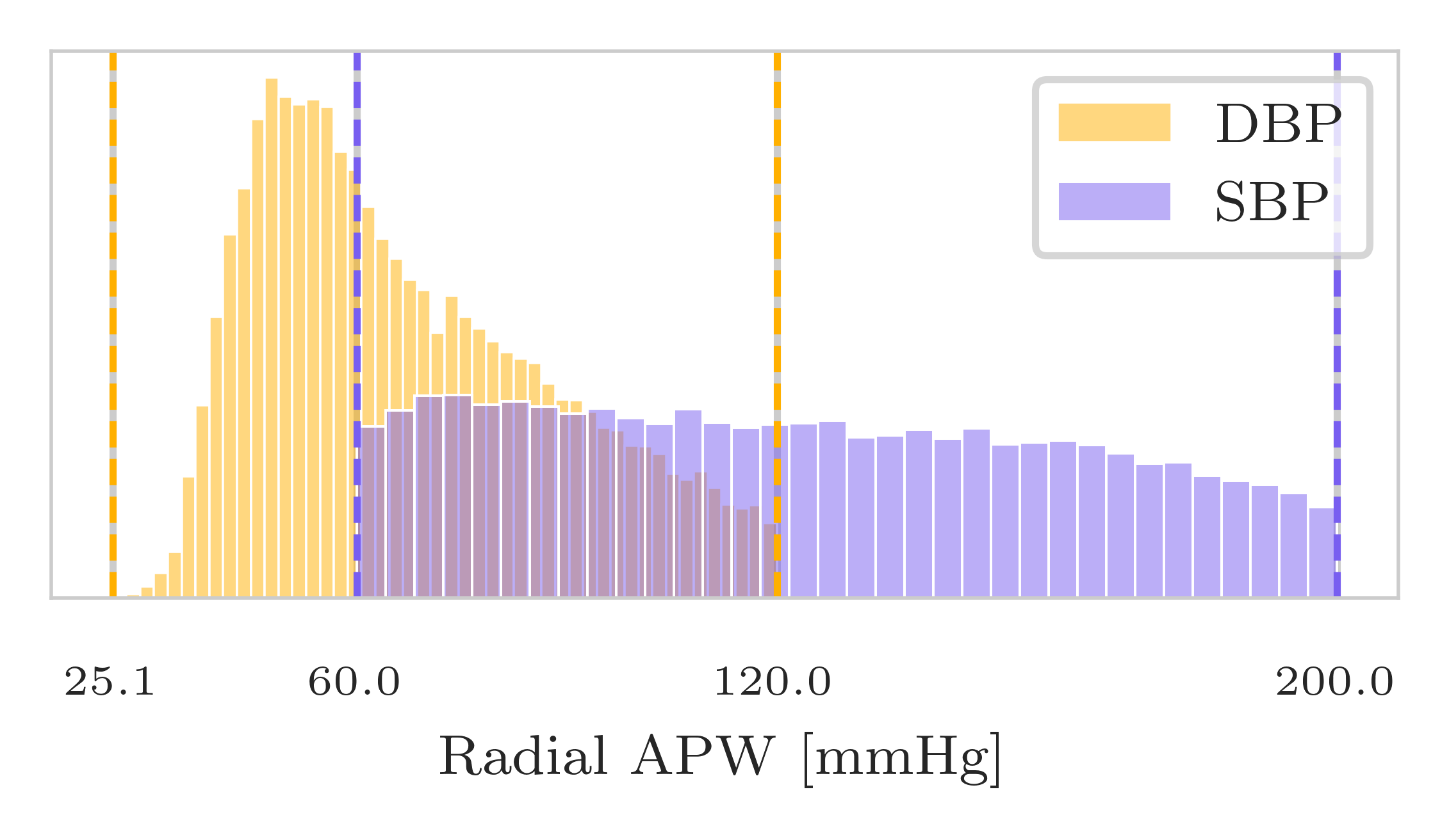}};
%Rounded Rect [id:dp21278909773161925] 
\draw  [color=black,draw opacity=.5 ][line width=.75]  (3,44.97) .. controls (3,32.29) and (13.29,22) .. (25.97,22) -- (633.03,22) .. controls (645.71,22) and (656,32.29) .. (656,44.97) -- (656,178.03) .. controls (656,190.71) and (645.71,201) .. (633.03,201) -- (25.97,201) .. controls (13.29,201) and (3,190.71) .. (3,178.03) -- cycle ;
%Rounded Rect [id:dp3311358031544489] 
\draw  [color=black,draw opacity=.5 ][line width=.75]  (2.5,255.36) .. controls (2.5,241.35) and (13.85,230) .. (27.86,230) -- (630.14,230) .. controls (644.15,230) and (655.5,241.35) .. (655.5,255.36) -- (655.5,558.64) .. controls (655.5,572.65) and (644.15,584) .. (630.14,584) -- (27.86,584) .. controls (13.85,584) and (2.5,572.65) .. (2.5,558.64) -- cycle ;
%Straight Lines [id:da7112717472919079] 
\draw [color={rgb, 255:red, 128; green, 128; blue, 128 }  ,draw opacity=1 ] [dash pattern={on 4.5pt off 4.5pt}]  (330,239) -- (329.75,578) ;
%Straight Lines [id:da1712245440175596] 
\draw [color={rgb, 255:red, 128; green, 128; blue, 128 }  ,draw opacity=1 ] [dash pattern={on 4.5pt off 4.5pt}]  (10.75,405) -- (646.75,405) ;
%Straight Lines [id:da29829025076660676] 
\draw [color={rgb, 255:red, 128; green, 128; blue, 128 }  ,draw opacity=1 ] [dash pattern={on 4.5pt off 4.5pt}]  (329.5,28) -- (329.5,195) ;

% Text Node
\draw (18,31) node [anchor=north west][inner sep=0.75pt]  [font=\small,color={rgb, 255:red, 0; green, 0; blue, 0 }  ,opacity=1 ] [align=left] {1) Sample Cardiac Parameters};
% Text Node
\draw (18,239) node [anchor=north west][inner sep=0.75pt]  [font=\small,color={rgb, 255:red, 0; green, 0; blue, 0 }  ,opacity=1 ] [align=left] {3) Remove Outliers};
% Text Node
\draw (344,239) node [anchor=north west][inner sep=0.75pt]  [font=\small,color={rgb, 255:red, 0; green, 0; blue, 0 }  ,opacity=1 ] [align=left] {4) Stack Single-Beats};
% Text Node
\draw (18,415) node [anchor=north west][inner sep=0.75pt]  [font=\small,color={rgb, 255:red, 0; green, 0; blue, 0 }  ,opacity=1 ] [align=left] {5) Apply Noise Model};
% Text Node
\draw (344,415) node [anchor=north west][inner sep=0.75pt]  [font=\small,color={rgb, 255:red, 0; green, 0; blue, 0 }  ,opacity=1 ] [align=left] {6) Bandpass Filter \& Normalization};
% Text Node
\draw (344,31) node [anchor=north west][inner sep=0.75pt]  [font=\small,color={rgb, 255:red, 0; green, 0; blue, 0 }  ,opacity=1 ] [align=left] {2) Simulate Hemodynamics};
% Text Node
\draw (24.75,-1) node [anchor=north west][inner sep=0.75pt]  [font=\large,color={black}  ,opacity=1 ] [align=left] {In-vivo Dataset Generation:};
% Text Node
\draw (24,207) node [anchor=north west][inner sep=0.75pt]  [font=\large,color={black}  ,opacity=1 ] [align=left] {Preprocessing:};

\end{tikzpicture}

}

%% file: tex/conclusion.tex
\section{Conclusion}

% We have demonstrated that SBI enables fine-grained uncertainty analysis and comparison of the relationship between various biomarkers and biosignals. 
In this paper, we have introduced a simulation-based inference methodology to analyze complex CV simulators and integrate them into clinical practices.
Our proposed framework performs statistical inference over critical cardiac parameters of a black-box CV simulator and provides a consistent and multi-dimensional quantification of uncertainty for individual measurements.
This uncertainty representation enabled us to identify an appropriate measurement model, recognize ambiguous inverse solutions, study the heterogeneity of sensitivity in the population considered, and understand dependencies between biomarkers in the inverse problem. 
Supported by in-vivo results, we empirically proved that simulation-based inference applied on 1D CV simulators can be employed to successfully track the temporal trend of critical biomarkers and detect low-SNR measurements. We have also discussed the challenge of model misspecification in scientific inquiry and proposed a hybrid learning strategy to overcome it in the context of full-body hemodynamics. 
In summary, 
% simulation-based inference enables scientists to address inverse problems in CV models, accounting for complex forward model dynamics and individualized uncertainty. 
our work provides foundations for a more effective use of CV simulations for cardiovascular research and personalized medicine.

% In summary, simulation-based inference allows scientists to study inverse problems arising from CV models with a methodology that acknowledges the forward model complexity and enables an individualised representation of uncertainty. These two aspects are necessary to fully leverage CV simulations for scientific enquiry and personalised medicine.

% facilitates iterations over the scientific loop and, thus, the discovery of real-world insights from CV simulations.
% The rigour and versatility of the SBI framework should eventually enhance the scientific and real-world impact of complex cardiovascular simulations.

%% file: tex/appendix.tex
\begin{appendices}
\section{Uncertainty analysis with SBI.} \label{sec:metrics}
Uncertainty analysis~\citep{sacks1989design, hespanhol2019understanding} regards identifiability as a continuous attribute of a model which allows ranking models by how much information the modeled observation process carries about the parameter of interest. We move away from the classical notion of statistical identifiability -- convergence in probability of the maximum likelihood estimator to the actual parameter value -- because this binary notion is not always relevant in practice and mainly applies to studies in the large sample size regime. In contrast, uncertainty analysis directly relates to the mutual information between the parameter of interest and the observation as expressed by the model considered. It captures that biased or noisy estimators are informative and may suffice for downstream tasks. 

As is standard in Bayesian uncertainty analyses, we look at credible regions $\Phi_\alpha(\mathbf{x})$ at different levels $\alpha$, which are directly extracted from the posterior distribution $p(\phi \mid \mathbf{x})$. Formally, a credible region is a subset, $\Phi_\alpha$, of the parameter space $\Phi$ over which the conditional density integrates to $\alpha$, i.e., $\Phi_\alpha: \int_{\phi \in \Phi_\alpha(\mathbf{x})} p(\phi \mid \mathbf{x}) \text{d}\phi = \alpha, \Phi_\alpha \subseteq \Phi$. In this paper, we consider the smallest covering union of regions, denoted by $\Phi_\alpha$, which is always unique in our case and in most practical settings. 

\subsection{Size of credible intervals (SCI).}
We rely on the SCI to shed light on the uncertainty of a parameter given a measurement process. 
The SCI at a level $\alpha$ is the expected size of the credible region at this level: $\mathbb{E}_{\mathbf{x}}[ \| \tilde{\Phi}_{\alpha}(\mathbf{x}) \| ]$, where $\| \cdot \|$ measures the size of a subset of the parameter space. In practice, we split the parameter space into evenly sized cells and count the number of cells belonging to the credible interval. As discussed in \appref{app:MI_identifiability}, there exists a relationship between SCI and mutual information~(MI). However, SCI is easier to interpret for domain experts than MI, as the former is expressed in the parameter's units. In addition, SCI is robust to multi-modality in contrast to point-estimator-based metrics (e.g., mean squared/absolute error) that cannot discriminate between two posterior distributions if they lead to the same point estimate. 

Algorithm~\ref{algorithm:average_credible_interval} describes a procedure to compute the size of credible intervals. In our experiments, we consider each dimension independently and discretize the space of value in $100$ cells. We finally report the average number of cells multiplied by the size of one cell in the physical unit of the parameter.

\begin{algorithm}[ht!]
\caption{Compute Average Size of Credible Intervals}
\label{algorithm:average_credible_interval}
\textbf{Input:} Dataset of observations $x$, Posterior distribution $p(\phi|x)$, Credibility level $\alpha$
\textbf{Output:} Average size of credible intervals

\begin{algorithmic}[1]
\STATE Initialize an empty list $CredIntSizes$
\FOR{each observation $x$ in the dataset}
    \STATE Generate samples from the posterior distribution: $\phi_{\text{samples}} =$ SampleFromPosterior($p(\phi | x)$)
    \STATE Discretize the parameter space into cells
    \STATE Initialize an empty list $CellCounts$
    \FOR{each sample $\phi$ in $\phi_{\text{samples}}$}
        \STATE Increase by one the count the cell covering $\phi$
    \ENDFOR
    \STATE Sort the $CellCounts$ list in descending order
    \STATE Append the minimum number of cells required to reach the credible level $\alpha$ to $CredIntSizes$.
\ENDFOR
\STATE Compute the average size of credible intervals by taking the mean of the $CredIntSizes$ list

\end{algorithmic}
\textbf{Return:} Average size of credible intervals
\end{algorithm}
\subsection{Mutual information and SCI}\label{app:MI_identifiability}
\begin{figure}
    \centering
    \includegraphics[width=.8\textwidth]{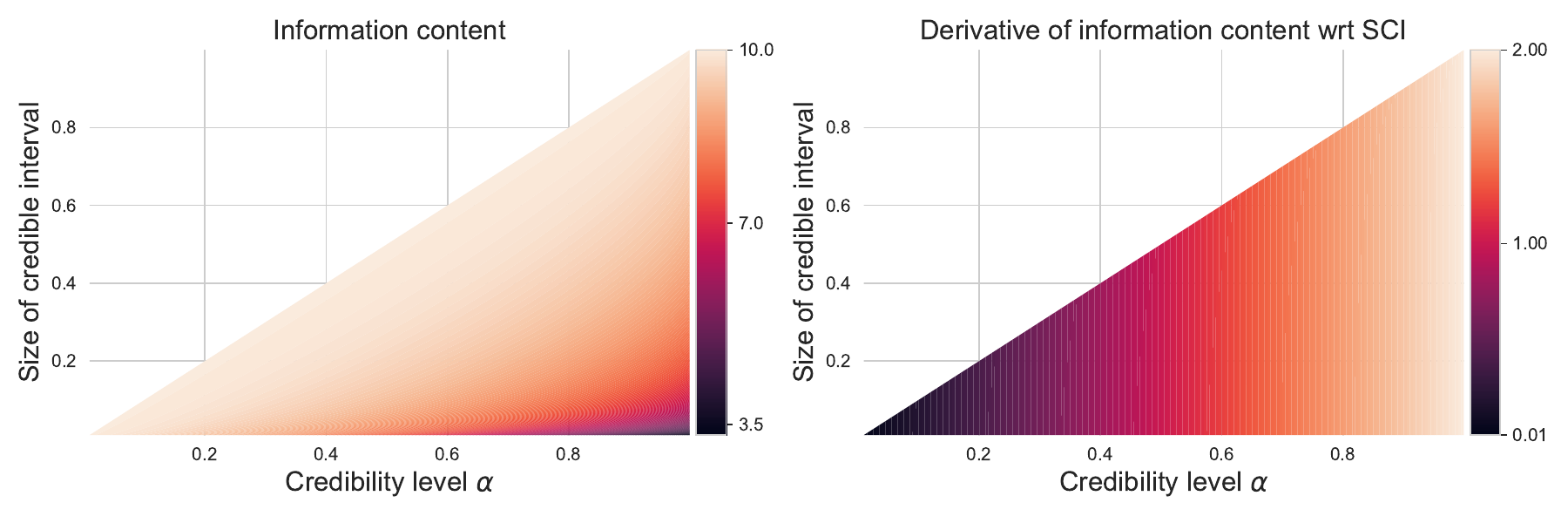}
    \caption{Relationship between the size of credible intervals and the information content present in the signal, for all credibility level. On the left: the plot of \eqref{eq:bits_SCI}; on the right: the plot of the derivative of \eqref{eq:bits_SCI} with respect to the SCI. We observe that larger SCI corresponds to larger number of bits required to encode the true value of the parameter of interest given the observation.}
    \label{fig:bits_SCI}
\end{figure}
In our experiments, we gave the average size of credible (SCI) intervals rather than the mutual information, as the former quantity is expressed in physical units that have a direct interpretation to specialists. We now discuss how the SCI relates to mutual information.

We aim to drive this discussion in the context of comparing the quality of two distinct measurement processes for inferring one quantity of interest. Formally, we denote these two measurements by $\mathbf{x}_1 \in \mathcal{X}_1$ and $\mathbf{x}_2 \in \mathcal{X}_2$ and the quantity of interest by $\phi \in \Phi$. 

Assuming a fixed marginal distribution $p(\phi)$ over the parameter, the two measurement processes $p(\mathbf{x}_1\mid \phi)$ and $p(\mathbf{x}_2 \mid \phi) $ define two joint distributions $p(
\mathbf{x}_1, \phi)$ and $p(\mathbf{x}_2, \phi) $. Considering a discretized space of parameters $\phi$, the mutual information can be written as 
\begin{align}
    \mathcal{I}(\phi, \mathbf{x}_i) = \mathcal{H}(\phi) - \mathcal{H}(\phi \mid \mathbf{x}_i),
\end{align}
 where $\mathcal{I}(\phi, \mathbf{x}_i)$ is the mutual information and $\mathcal{H}$ the entropy. As the marginal entropy of the parameter remains constant, it is clear that only the second term matters for comparing the information content of the two measurement processes. The quantity $\mathcal{H}(\phi \mid \mathbf{x}_i)$ can be interpreted as the average remaining number of bits necessary to encode $\phi$ if we know $\mathbf{x}_i$. The average is taken over the joint distribution induced by the marginal distribution of $\phi$ and the measurement process $p(\mathbf{x}_i\mid \phi)$.

From this interpretation, choosing the one with the highest mutual information is a well-motivated criterion for choosing between two measurement processes. Said differently, we are looking for measurement processes with the smallest $\mathcal{H}(\phi \mid \mathbf{x}_i)$, the one leading to small uncertainty about the correct value of $\phi$.

We use an information theory point of view to explain why, similarly to mutual information maximization~\citep{gao2020reducing}, aiming for the measurement process with the smallest SCI is a sound approach. Let us consider the measurement process $p(\mathbf{x}_i \mid 
\phi)$ leading credible intervals with size $S(\alpha, \mathbf{x}_i)$ for a certain credibility level $\alpha$ and observation $\mathbf{x}_i$. Similarly to what we do in practice to compute the SCI, we discretize the space of parameters into $N$ cells. The SCI is then defined as the minimal number of cells required to cover the credible region at level $\alpha$. From the SCI, we can say that the true value of $\phi$ belongs to one of the $S(\alpha, \mathbf{x}_i)$ cells of the credible region with probability $\alpha$ or to one of the $N - S(\alpha, \mathbf{x}_i)$ remaining cells with a probability $1-\alpha$. From this, we can bound the average number of bits required to encode the true value of $\phi$ given the observation $\mathbf{x}_i$ as a function of the SCI $S$ and the credibility level $\alpha$ as
\begin{align}
    \text{N bits} \leq -\alpha  \log_2\frac{\alpha}{S(\alpha, \mathbf{x}_i)} - (1-\alpha) \log_2 \frac{1-\alpha}{N - S(\alpha, \mathbf{x}_i)}. \label{eq:bits_SCI}
\end{align}

\figref{fig:bits_SCI} shows the relationship between this bound and the credibility level $\alpha$ and SCI. We treat SCI and the credibility $\alpha$ as independent quantities, as different measurement processes can lead to different relationships between these two quantities. We must notice that given a credibility level $\alpha$, smaller $SCI$ corresponds to better bounds. We can conclude that selecting models with the smallest SCI for a given credibility level is a sound approach with a similar interpretation as making this choice based on mutual information.

\subsection{Average Coverage AUC.}
Given samples from the joint distribution $p(\phi, \mathbf{x})$, credible intervals are expected to contain the true value of the parameter at a frequency equal to the credibility level $\alpha$, that is, $\mathbb{E}_{p(\phi, \mathbf{x})} \left[1_{\Phi_\alpha}(\phi) \right] = \alpha $, where $1$ is the indicator function. In this work, we do not have access to the true posterior but a surrogate $\tilde{p}$ of it. Hence, the coverage property of credible regions, which support the interpretation of uncertainty, may be violated, even when the forward model and the prior accurately describe the data. The ACAUC $C(\tilde{p}(\phi \mid \mathbf{x}), \mathcal{D})$ of a surrogate posterior $\tilde{p}$ is a metric, computed on a set $\mathcal{D}:=\{(\phi_j^\star, \mathbf{x_j})\}_{j=1}^N$, that measures whether the surrogate's credible regions respect coverage. We compute the ACAUC as
\begin{align}
    C(\tilde{p}(\phi \mid \mathbf{x}), \mathcal{D}) &= \frac{1}{k}\sum_{i=1}^k \bigg| \frac{i}{k} - \frac{1}{N}\sum_{j=1}^N  1_{\tilde{\Phi}_{\frac{i}{k}}(\mathbf{x}_j)}(\phi^\star_j) \bigg| \\
&\approx \mathbb{E}_{\alpha \sim \mathcal{U}(]0, 1])}\left[ \bigg| \alpha - \mathbb{E}_{p(\phi^\star, \mathbf{x})}\left[ 1_{\tilde{\Phi}_{\alpha}(\mathbf{x})}(\phi^\star) \right]  \bigg| \right],
\end{align}
where $\tilde{\Phi}_{\frac{i}{k}}(\mathbf{x}_j)$ is the credible region at level $\alpha=\frac{i}{k}$ corresponding to the surrogate posterior distribution $\tilde{p}(\phi \mid \mathbf{x}_j)$, and we consider a large test set made of $N$ pairs $(\phi^\star_j, \mathbf{x}_j)$.
The ACAUC directly relates to how much the surrogate posterior model violates the coverage property over all possible levels $\alpha \in \left] 0, 1 \right]$. 

Algorithm~\ref{algorithm:statistical_calibration} returns the distribution of minimum credibility levels required to not reject the true value of $\phi$. Under calibration, these values should be uniformly distributed -- we expect to reject falsely the true value with a frequency equal to the credibility level chosen. We report the integral (along credibility levels $\alpha$) between the observed cumulative distribution function~(CDF) of minimum credibility levels and the CDF of a uniform distribution. This metric equals $0$ under perfect calibration and is bounded by $0.5$. We report the calibration for each dimension independently as the metric does not generalize to multiple dimension.

\begin{algorithm}[ht!]
\caption{Statistical Calibration of Posterior Distribution}
\label{algorithm:statistical_calibration}
\textbf{Input:} Dataset of pairs $\mathcal{D} = \{(\phi_i, x_i)\}$, Posterior distribution $p(\phi|x)$, Number of samples $N$.\\
\textbf{Output:} Distribution of minimum credibility levels.

\begin{algorithmic}[1]
\STATE Initialize an empty list $CredLevels$
\FOR{$(\phi_i, x_i) \in \mathcal{D}$}
    \STATE Initialize an empty list $Samples$
    \FOR{$i = 1$ to $N$}
        \STATE Sample $\phi_i$ from $p(\phi|x)$
        \STATE Append $\phi_i$ to $Samples$
    \ENDFOR
    \STATE Sort $Samples$
    \STATE Compute the rank (position in ascending order) $r$ of $\phi$ in $Samples$
    \STATE Set $CredLevel = \frac{r}{N}$
    \STATE Append $CredLevel$ to $CredLevels$
\ENDFOR
% \STATE Compute the distribution of minimum credibility levels based on $CredLevels$
% \RETURN Distribution of minimum credibility levels
\end{algorithmic}
\textbf{Return:} $CredLevels$.
\end{algorithm}

\section{Supplementary results}\label{app:supp_results}

\begin{figure}[H]
    \centering
    \includegraphics[width=1.\textwidth]{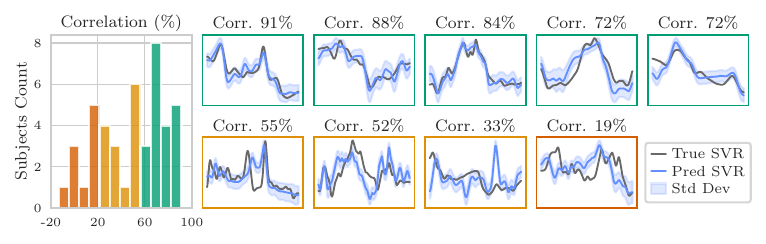}
    \caption{Left: histogram of the per-patient Spearman correlations between the measured SVR in VitalDB and the predicted SVR by NPE, which is trained solely on simulated APWs. Right: we plot the normalized SVR tracking provided by VitalDB (SVR) together with the NPE prediction (Pred SVR) for patients with different correlations.
    }
    \label{fig:tracking_vitaldb_svr}
\end{figure}

\section{Datasets}~\label{app:dataset}
In this section we provide further details and examples of both the in-silico and in-vivo datasets. Specifically, we list the complete set of free parameters considered in the whole-body hemodynamics model. We then provide examples of in-silico APWs and PPGs at different noise levels. Finally, we elucidate the pre-processing steps taken for the VitalDB datasets. We observe that real-world data contains (a) degenerated beats and (b) different physiological parameter dynamics accross beats. These observations motivate the use of the stochastic noise model in the in-silico dataset.

\subsection{In-silico Dataset}
\subsubsection{Whole-body Hemodynamics Model Parameterization}\label{app:hemodynamics_model}
To construct the in-silico dataset described in \autoref{sec:simulations}, we consider a population aged $25$ to $75$, where the following free parameters, which model heterogeneous cardiac, arterial, and vascular bed properties, 
vary from one simulation to the other:
\begin{itemize}
    \item Heart function:
    \begin{itemize}
        \item Heart Rate~(HR).
        \item Stroke Volume~(SV).
        \item Left Ventricular Ejection Time~(LVET). \textit{Note:} LVET changes as a deterministic function of HR and SV.
        \item Peak Flow Time~(PFT).
        \item Reflected Fraction Volume~(RFV).
    \end{itemize}
    \item Arterial properties:
    \begin{itemize}
        \item $Eh = R_{d} (k_1 e^{k_2 R_d} + k_3)$ where $k1, k2$ are constant and $k3$ follows a deterministic function of age.
        \item Length of proximal aorta
        \item Diameter of larger arteries
    \end{itemize}
    \item Vascular beds
    \begin{itemize}
        \item Resistance adjusted to achieve mean average pressure~(MAP) distribution compatible with real-world studies.
        \item Compliance adjusted to achieve realistic peripheral vascular compliance~(PVC) compatible with real-world studies.
    \end{itemize}
\end{itemize}
The interested reader will find further details in \cite{charlton2019modeling}.

\subsubsection{Measurement model}
\begin{figure}[ht]
    % First Row

\begin{subfigure}[b]{0.3\textwidth}\includegraphics[width=1.\textwidth]{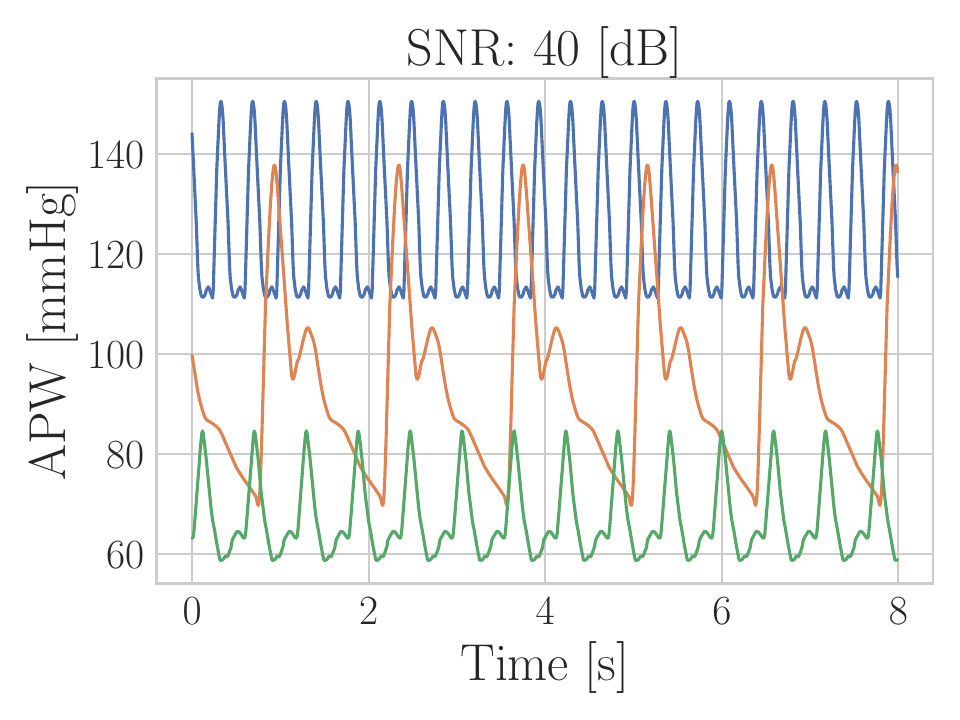}
    \caption{}\label{fig:noise_level_0_apw}
    % \vspace{-1.5em}
\end{subfigure}
\begin{subfigure}[b]{0.3\textwidth}\includegraphics[width=1.\textwidth]{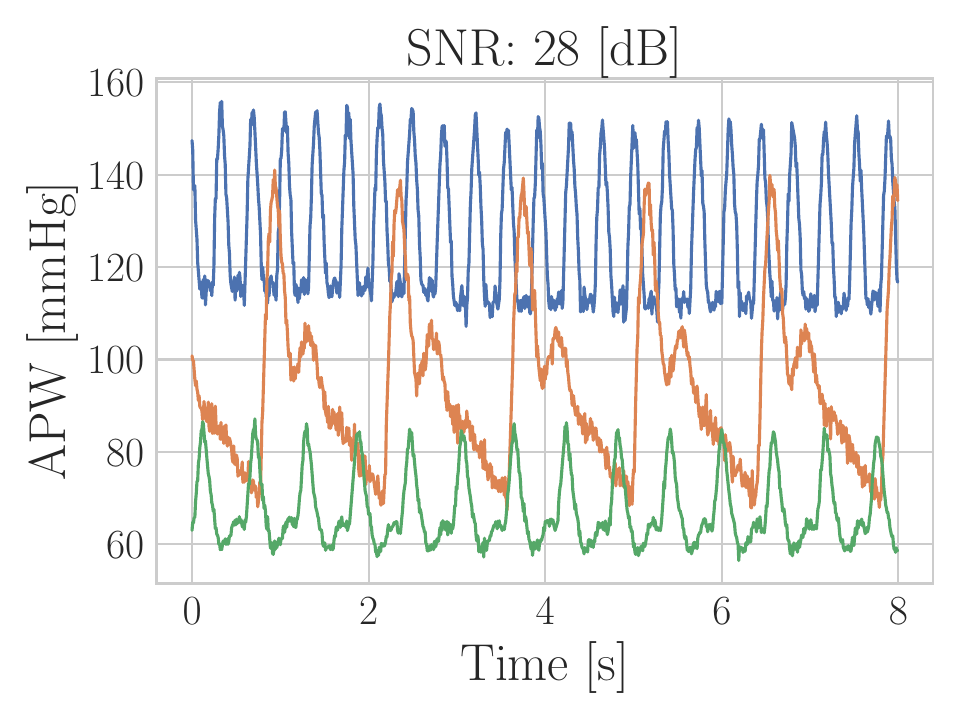}
    \caption{}\label{fig:noise_level_1_apw}
    % \vspace{-1.5em}
\end{subfigure}
\begin{subfigure}[b]{0.3\textwidth}\includegraphics[width=1.\textwidth]{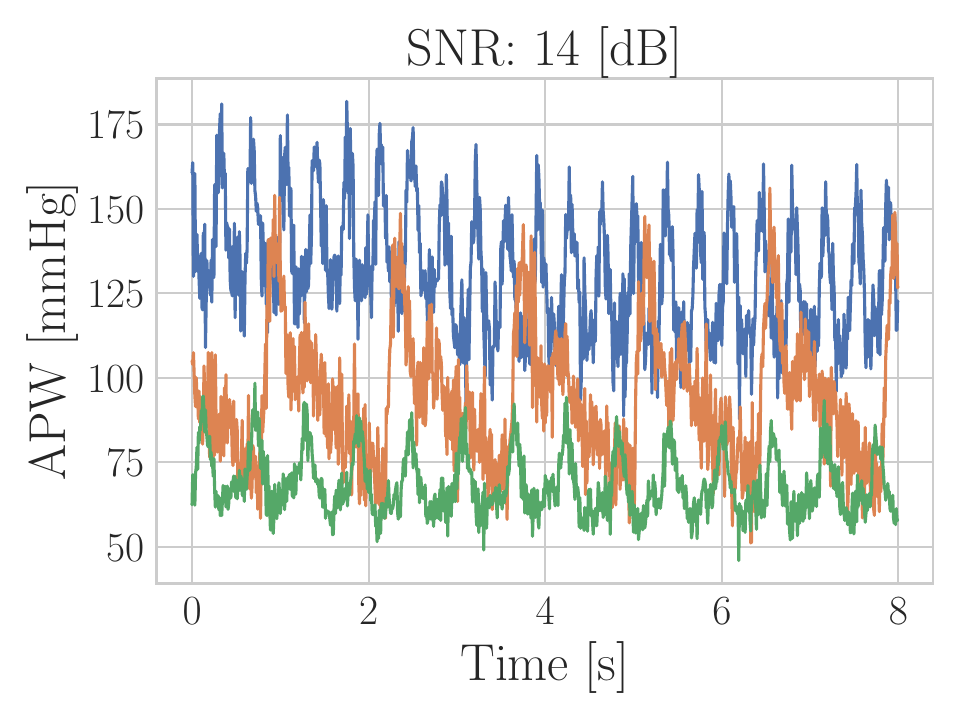}
    \caption{}\label{fig:noise_level_2_apw}
    % \vspace{-1.5em}
\end{subfigure}

\begin{subfigure}[b]{0.3\textwidth}\includegraphics[width=1.\textwidth]{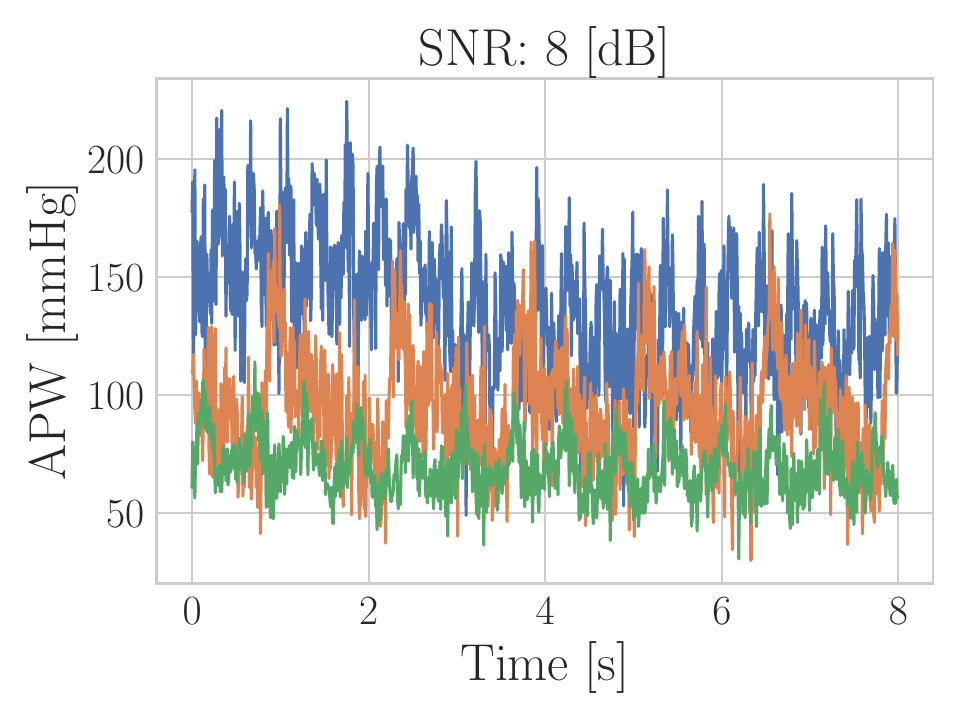}
    \caption{}\label{fig:noise_level_3_apw}
    % \vspace{-1.5em}
\end{subfigure}
\begin{subfigure}[b]{0.3\textwidth}\includegraphics[width=1.\textwidth]{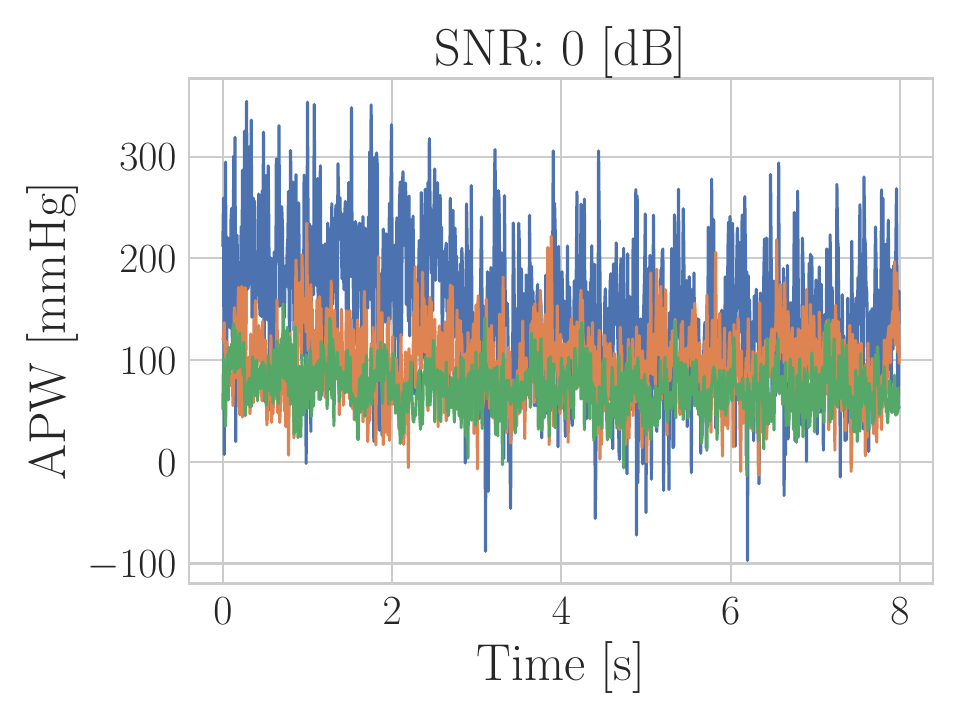}
    \caption{}\label{fig:noise_level_4_apw}
    % \vspace{-1.5em}
\end{subfigure}
\begin{subfigure}[b]{0.3\textwidth}\includegraphics[width=1.\textwidth]{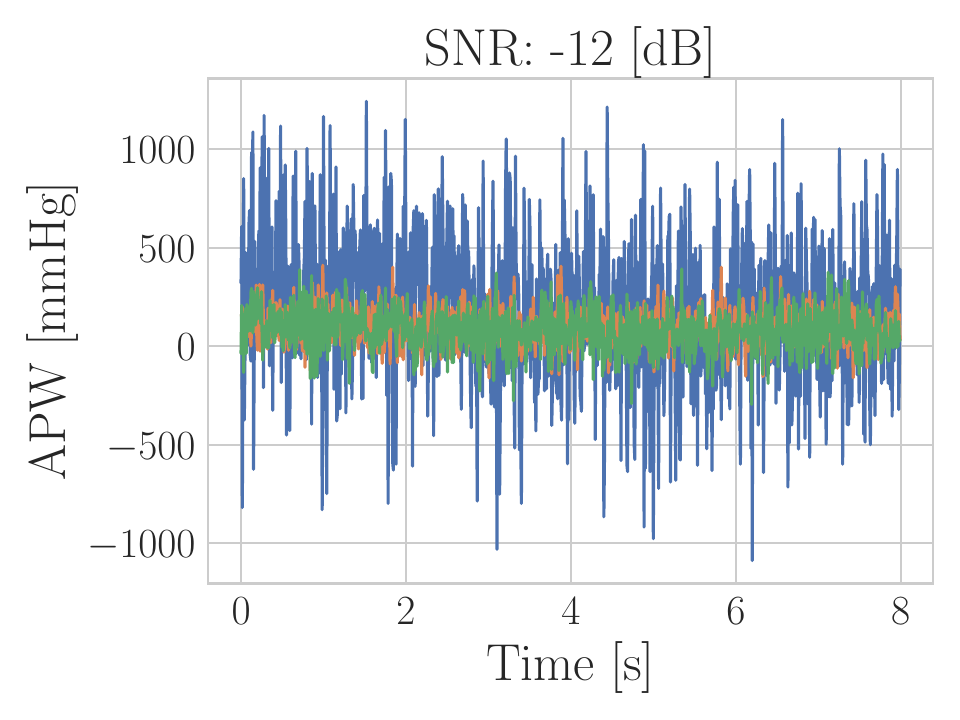}
    \caption{}\label{fig:noise_level_5_apw}
    % \vspace{-1.5em}
\end{subfigure}
\caption{Three randomly picked synthetic APWs at increasing levels of noise.}\label{fig:APW_gen_at_several_noise}
\end{figure}

To bridge the gap between simulated and real-world data, we apply a stochastic measurement model detailed in \autoref{sec:simulations}. To better understand the effect of the noise, we show examples of APWs at different noise levels in \autoref{fig:APW_gen_at_several_noise}.

\begin{figure}[H]
\centering
  \includegraphics[width=1.\linewidth]{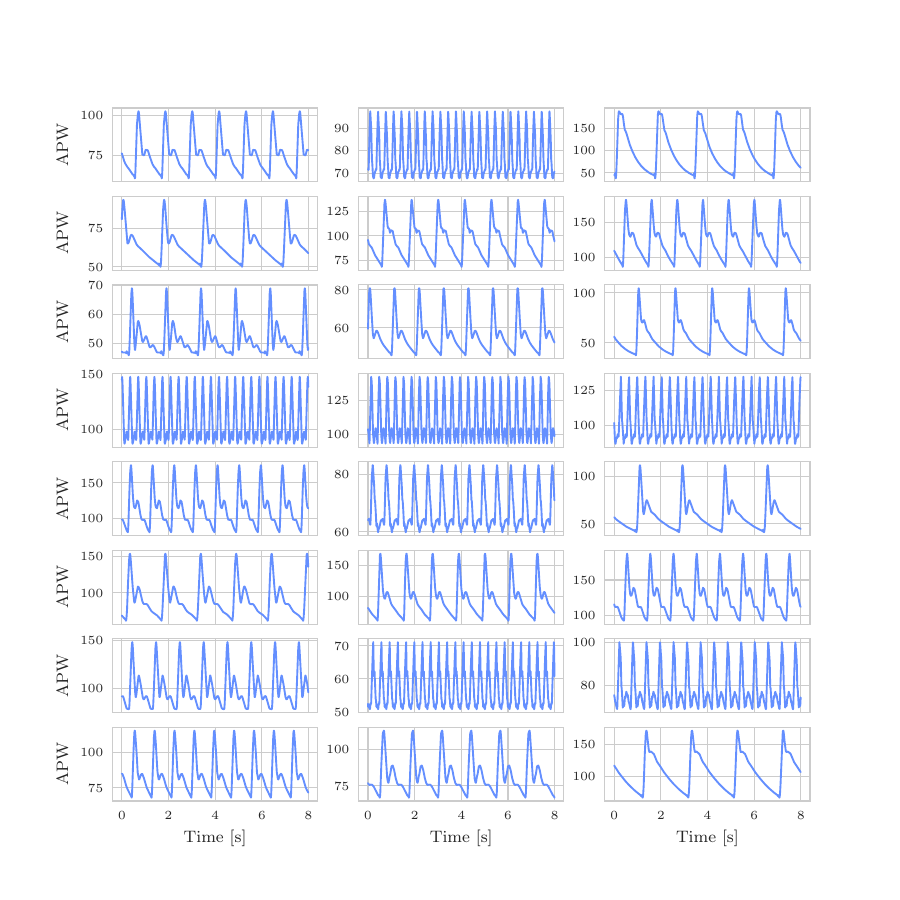}
  \caption{Random examples of synthetic APWs at the radial artery without the stochastic noise model.}
  \label{app:fig:synthetic_apw}
\end{figure}

\begin{figure}[H]
\centering
  \includegraphics[width=1.\linewidth]{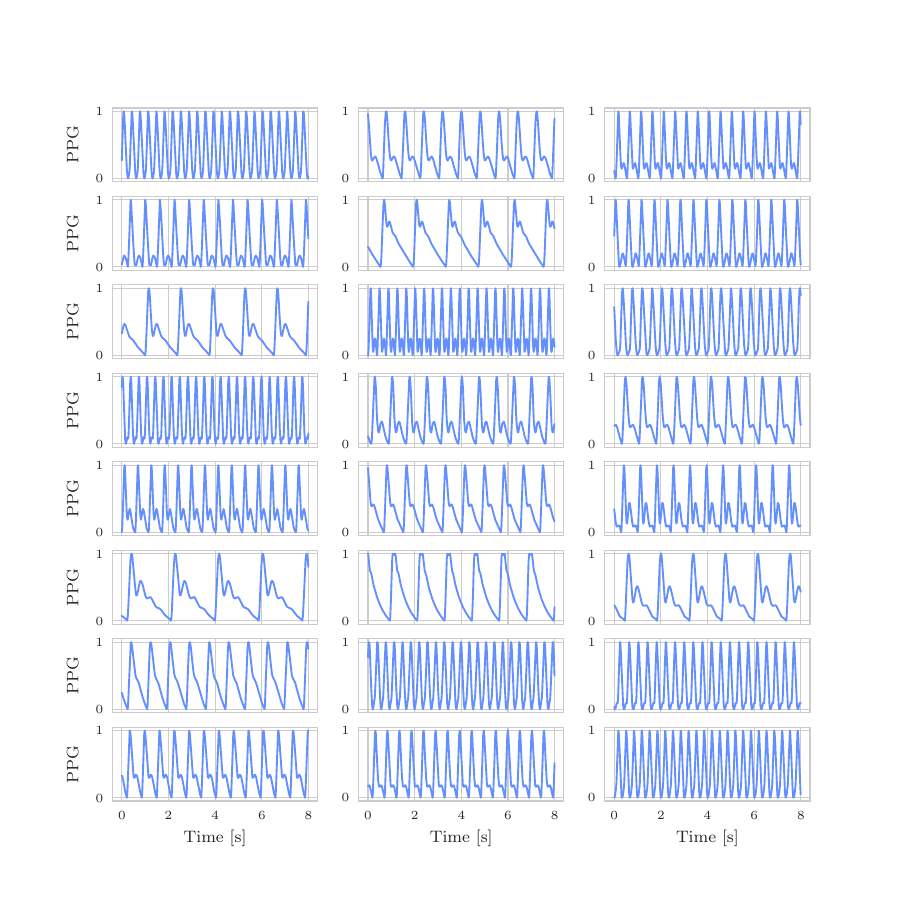}
  \caption{Random examples of synthetic digital PPGs without the stochastic noise model.}
  \label{app:fig:synthetic_ppg}
\end{figure}

\begin{figure}[H]
\centering
  \includegraphics[width=1.\linewidth]{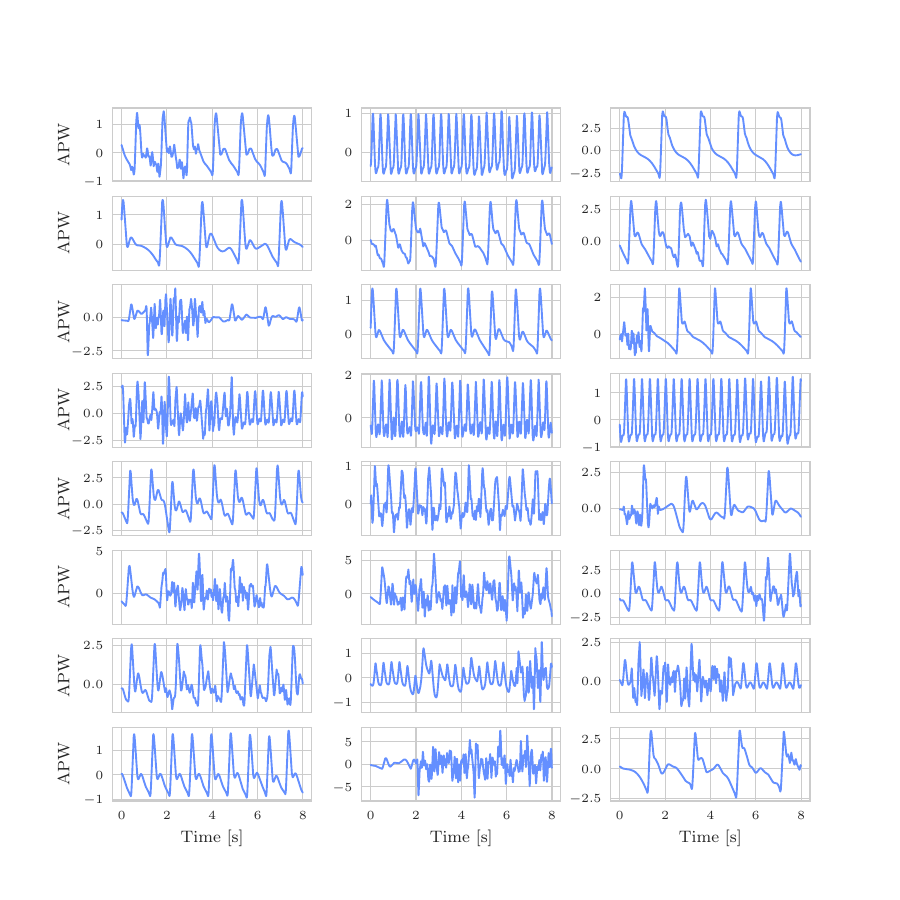}
  \caption{Random examples of synthetic APWs after adding the stochastic noise model and the Butterworth second-order bandpass filter.}
  \label{app:fig:synthetic_apw_preprocessed}
\end{figure}

\begin{figure}[H]
\centering
  \includegraphics[width=1.\linewidth]{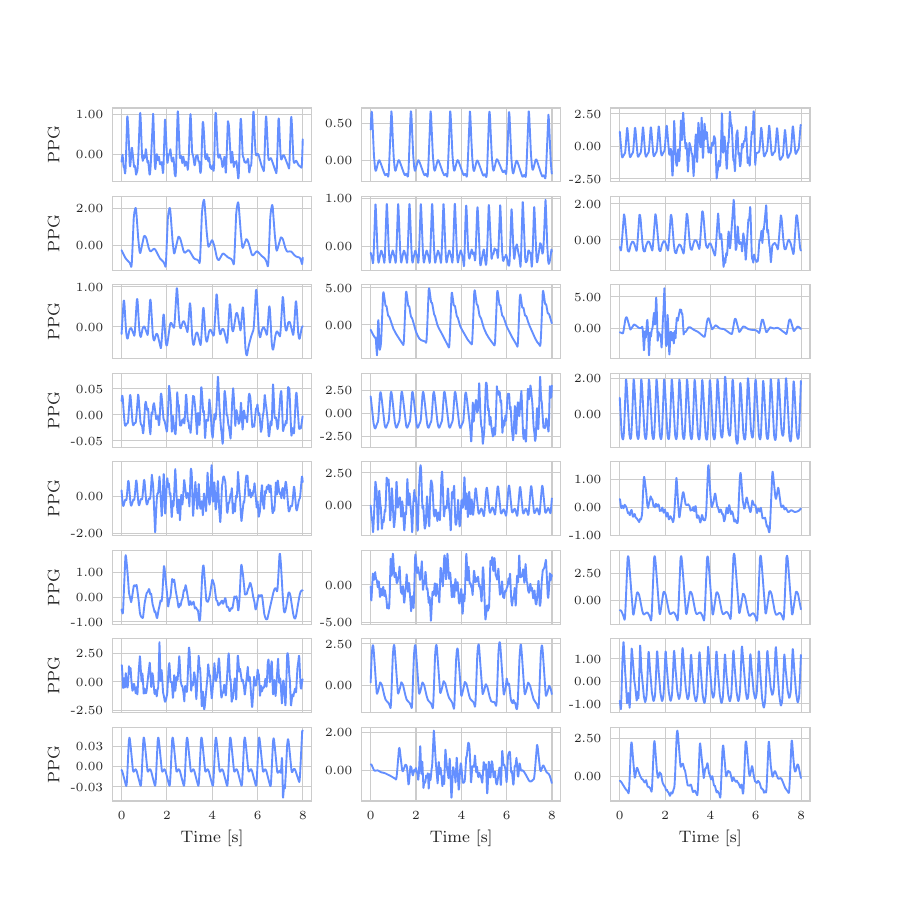}
  \caption{Random examples of synthetic digital PPGs after adding the stochastic noise model and the Butterworth second-order bandpass filter.}
  \label{app:fig:synthetic_ppg_preprocessed}
\end{figure}

The resulting  APWs at the radial artery before and after applying the stochastic noise (with a Butterworth second-order bandpass filter) are showcased in \autoref{app:fig:synthetic_apw} and \autoref{app:fig:synthetic_apw_preprocessed} respectively. While the corresponding digital PPGs are showcased in \autoref{app:fig:synthetic_ppg} and \autoref{app:fig:synthetic_ppg_preprocessed} 

% \begin{figure}[h]
%     \centering
%     \includegraphics[width=.75\textwidth]{PNAS/figures/pulsedb_mimic_quicklook.pdf}
%     \caption{Waveforms reproduced from the MIMIC-III waveform database~\citep{moody2020mimic} subset of the PulseDB dataset~\citep{wang2023pulsedb}.}
%     \label{fig:MIMIC}
% \end{figure}
\subsection{VitalDB Dataset} 
\label{app:vitaldb}
\begin{figure}[H]
\centering
  \includegraphics[width=.9\linewidth]{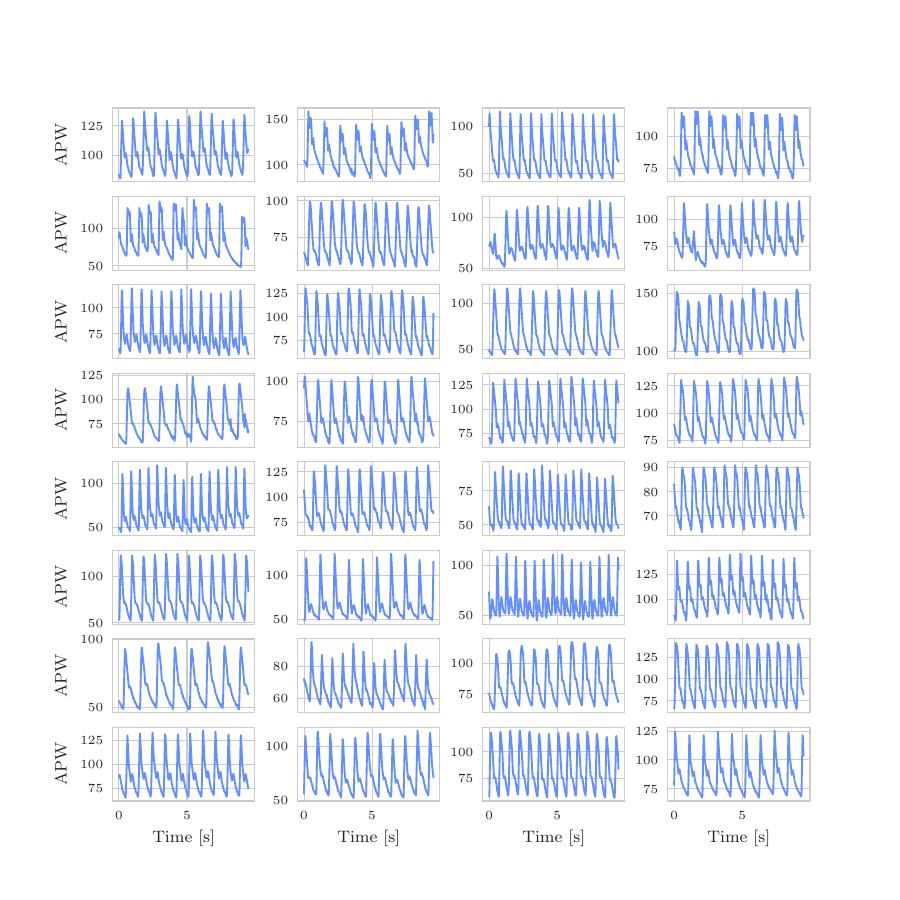}
  \caption{}
  \label{app:fig:vitaldb_apw}
\end{figure}

\begin{figure}[H]
\centering
  \includegraphics[width=.9\linewidth]{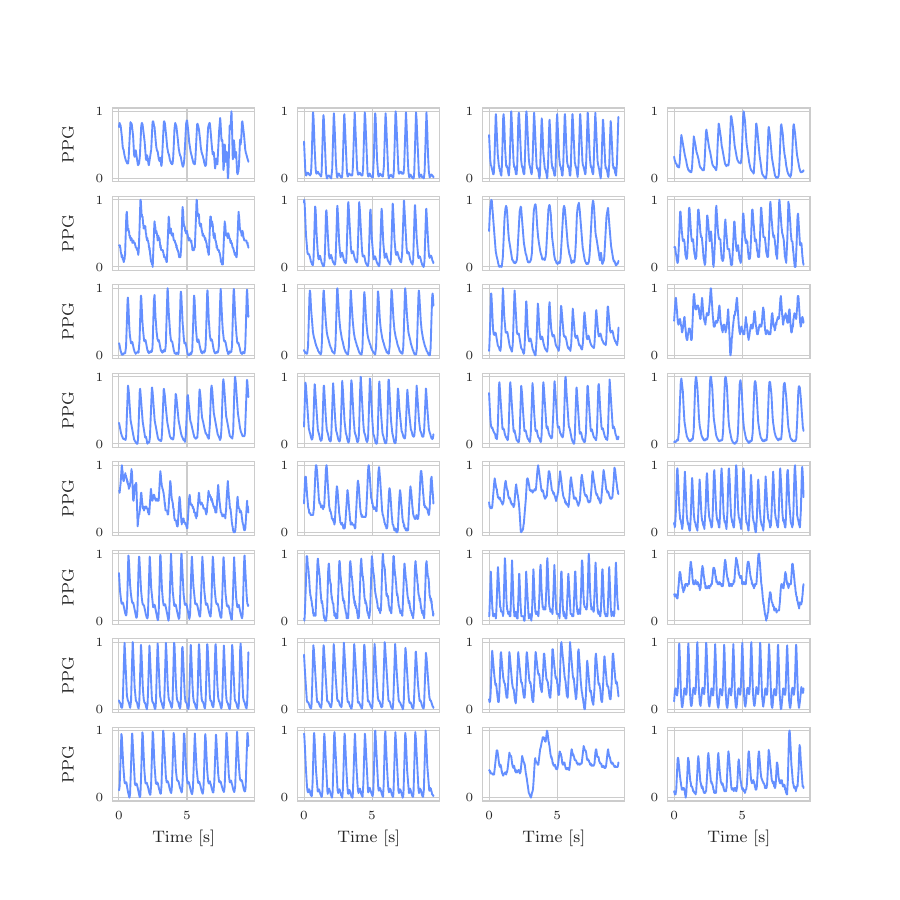}
  \caption{}
  \label{app:fig:vitaldb_ppg}
\end{figure}

%Real-world experiments are crucial for assessing the potential misspecifications of models developed in silico. While SBI can provide valuable insights and robust predictions, it is only through comparison with real-world data that we can validate these models and ensure their applicability to clinical settings. This section focuses on the application of the proposed SBI framework to predict heart rate (HR), cardiac output (CO), and systemic vascular resistance (SVR) from biosignals taken from the VitalDB dataset \cite{Lee2022VitalDBAH}.
% Train:
% Total number of measurements: 25086
% Number of unique caseid values: 148
% Number of items per caseid: 169.5
% Test:
% Total number of measurements: 10348
% Number of unique caseid values: 49
% Number of items per caseid: 211.18367346938774
VitalDB is a comprehensive, publicly available dataset that includes a wide range of physiological signals and clinical data collected from non-cardiac surgery patients who underwent routine or emergency surgery. 
In our analysis, we select APW and PPG recordings in which HR, CO and SVR measurements are available. We perform pre-processing steps to ensure the quality and reliability of the data, i.e., we filter out waveforms with missing values (NaNs), implausible minimum or maximum values, steep consecutive point changes, or a number of peaks inconsistent with the expected HR range. We divide the remaining waveforms into $8$-second segments to match the time resolution used in the in-silico simulations. We then process the signals in each segment using a Butterworth second-order bandpass filter with a frequency range of $0.5–10$ Hz to remove noise and artifacts.
Example of the resulting APWs and PPGs are shown in \autoref{app:fig:vitaldb_apw} and \autoref{app:fig:vitaldb_ppg}, respectively.

After pre-processing, we split the patients (and corresponding waveforms) into training, validation and test set. The training and validation set consist of $25,086$ waveforms from $148$ subjects, and the test set includes $10,348$ waveforms from $49$ subjects. 

\section{Normalizing flows} \label{app:NF}
We provide additional details on the normalizing flows used to model the posterior distributions. In all our experiments, we apply the same training and model selection procedures. Moreover we use the same neural network architecture for all experiments.

We rely on the open-source libraries PyTorch~\citep{NEURIPS2019_9015} and \href{https://github.com/AWehenkel/Normalizing-Flows}{Normalizing Flows}, a lightweight library to build NFs built upon the abstraction of NFs as Bayesian networks from \citep{wehenkel2021graphical}. 

\subsection{Training setup}
We randomly divide the complete dataset into $70\%$ train, $10\%$ validation, and $20\%$ test sets. We optimize the parameters of the neural networks with stochastic gradient descent on \eqref{eq:loss} with the Adam optimizer~\citep{kingma2014adam}. We use a batch size equal to $100$, a fixed learning rate ($= 10^{-3}$), and a small weight decay ($= 10^{-6}$). We train each model for $500$ epochs and evaluate the validation loss after each epoch. The best model based on the lowest validation loss was returned and used to obtain the results presented in the paper. All data are normalized based on their standard deviation and mean on the training set. For the time series, we compute one value across the time dimension. 

\subsection{Neural network architecture}
We use the same neural network architecture for all the results reported. It is made of a $3$-step autoregressive normalizing flow~\citep{papamakarios2017masked} combined with a convolutional neural network~(CNN) to encode the $8$-second segments sampled at $125Hz$ ($\in \mathbb{R}^{1000}$). The CNN is made of the following layers:
\begin{enumerate}
    \item 1D Convolution with no padding, kernel size $= 3$, stride $= 2$, $40$ channels, and  followed by ReLU;
    \item 1D Convolution with no padding, kernel size $= 3$, stride $= 2$, $40$ channels, and  followed by ReLU;
    \item 1D Convolution with no padding, kernel size $= 3$, stride $= 2$, $40$ channels, and  followed by ReLU;
    \item Max pooling with a kernel $=3$;
    \item 1D Convolution with no padding, kernel size $= 3$, stride $= 2$, $20$ channels, and  followed by ReLU;
    \item 1D Convolution with no padding, kernel size $= 3$, stride $= 2$, $10$ channels, and  followed by ReLU,
\end{enumerate}
leading to a $90$ dimensional representation of the input time series. The $90$-dimensional embedding is concatenated to the $age$ and denoted $\mathbf{h}$. Then, $\mathbf{h}$ is passed to the NF as an additional input to the autoregressive conditioner~\citep{papamakarios2017masked, wehenkel2021graphical}. 

The NF is made of a first autoregressive step that inputs both the $91$ conditioning vector $\mathbf{h}$ and the parameter vector and outputs $2$ real values $\mu_i(\phi_{<i}, \mathbf{h}), \sigma_i(\phi_{<i}, \mathbf{h}) \in \mathbb{R}$ per parameter in an autoregressive fashion. Then the parameter vector is linearly transformed as $u_i = \phi_i e^{\sigma_i(\phi_{<i}, \mathbf{h})} + \mu_i(\phi_{<i}, \mathbf{h})$. The vector $\mathbf{u} := [u_1, \dots, u_k]$ is then shuffled and passed through 2 other similar transformations, leading to a vector denoted $\mathbf{z}$, which eventually follows a Gaussian distribution after learning~\citep{papamakarios2021normalizing}. The $3$ autoregressive networks have the same architecture: a simple masked multi-layer perceptron with ReLu activation functions and $3$ hidden layers with $350$ neurons each. We can easily compute the Jacobian determinant associated with such a sequence of autoregressive affine transformations on the vector $\phi$ and thus compute \eqref{eq:loss}. 

We can easily show that the Jacobian determinant is equal to the product of all scaling factors $e^{\sigma_i}$. We also directly see that ensuring these factors are strictly greater than $0$ enforce a continuously invertible Jacobian for all value of $\phi$ and thus continuous bijectivity of the associated transformation.

As mentioned, under perfect training, the mapping from $\Phi$ to $\mathcal{Z}$ defines a continuous bijective transformation that transforms samples from $\phi \sim p(\phi \mid \mathbf{h})$ into samples $\mathbf{z} \sim \mathcal{N}(0,I)$. As the transformation is bijective, we can sample from $p(\phi \mid \mathbf{h})$ by inverting the transformation onto samples from $\mathcal{N}(0,I)$. As the transformation is autoregressive, we can invert it by doing the inversion sequentially for all dimensions as detailed in \citep{papamakarios2021normalizing, wehenkel2021graphical, papamakarios2017masked}.

\end{appendices}